\documentclass{article}

\usepackage{arxiv}

\usepackage[utf8]{inputenc} 
\usepackage[T1]{fontenc}    
\usepackage{hyperref}       
\usepackage{url}            
\usepackage{booktabs}       
\usepackage{amsfonts}       
\usepackage{nicefrac}       
\usepackage{microtype}      
\usepackage{lipsum}		
\usepackage{graphicx}
\usepackage{doi}
\usepackage{dsfont}
\usepackage{amsmath}
\usepackage[caption=false,font=normalsize,labelfont=sf,textfont=sf]{subfig}
\newcommand{\ie}{\textit{i}.\textit{e}.}
\newcommand{\eg}{\textit{e}.\textit{g}.}

\title{Towards Experience Replay for Class-Incremental Learning in Fully-Binary Networks}

\author{ \href{https://orcid.org/0009-0000-5109-6602}{\includegraphics[scale=0.06]{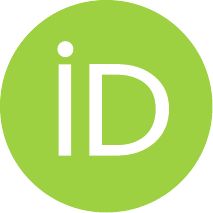}\hspace{1mm}Yanis BASSO-BERT}\\
	Univ. Grenoble Alpes, CEA, List\\
	F-38000 Grenoble\\
	France \\
	\texttt{yanis.bassobert@cea.fr} \\
	\And
	\href{https://orcid.org/0000-0003-2400-165X}{\includegraphics[scale=0.06]{orcid.pdf}\hspace{1mm}Anca MOLNOS}\\
	Univ. Grenoble Alpes, CEA, List\\
	F-38000 Grenoble\\
	France \\
	\texttt{anca.molnos@cea.fr} \\
	\And
	\href{https://orcid.org/0000-0002-7260-2786}{\includegraphics[scale=0.06]{orcid.pdf}\hspace{1mm}Romain LEMAIRE}\\
	Univ. Grenoble Alpes, CEA, List\\
	F-38000 Grenoble\\
	France \\
	\texttt{romain.lemaire@cea.fr} \\
	\And
	\href{https://orcid.org/0000-0001-8925-0441}{\includegraphics[scale=0.06]{orcid.pdf}\hspace{1mm}William GUICQUERO}\\
	Univ. Grenoble Alpes, CEA, Leti\\
	F-38000 Grenoble\\
	France \\
	\texttt{william.guicquero@cea.fr} \\
	\And
	\href{https://orcid.org/0000-0002-0145-2186}{\includegraphics[scale=0.06]{orcid.pdf}\hspace{1mm}Antoine DUPRET}\\
	Univ. Grenoble Alpes, CEA, Leti\\
	F-38000 Grenoble\\
	France \\
	\texttt{antoine.dupret@cea.fr} \\
}

\renewcommand{\shorttitle}{\textit{arXiv} Template}

\hypersetup{
pdftitle={Towards Experience Replay for Class-Incremental Learning in Fully-Binary Networks},
pdfsubject={cs.ai},
pdfauthor={Yanis BASSO-BERT, Anca MOLNOS, Romain LEMAIRE, William GUICQUERO, Antoine DUPRET},
pdfkeywords={Continual Learning, Class-Incremental Learning, Replay Methods, Binary Neural Network, Quantization},
}

\begin{document}
\maketitle

\begin{abstract}
Binary Neural Networks (BNNs) are a promising approach to enable Artificial Neural Network (ANN) implementation on ultra-low power edge devices. 
Such devices may compute data in highly dynamic environments, in which the classes targeted for inference can evolve or even novel classes may arise, requiring continual learning. 
Class Incremental Learning (CIL) is a common type of continual learning for classification problems, that has been scarcely addressed in the context of BNNs. Furthermore, most of existing BNNs models are not fully binary, as they require several real-valued network layers, at the input, the output, and for batch normalization. 
This paper goes a step further, enabling class incremental learning in Fully-Binarized NNs (FBNNs) through four main contributions. 
We firstly revisit the FBNN design and its training procedure that is suitable to CIL. 
Secondly, we explore loss balancing, a method to trade-off the performance of past and current classes.
Thirdly, we propose a semi-supervised method to pre-train the feature extractor of the FBNN for transferable representations. Fourthly, two conventional CIL methods, \ie, Latent and Native replay, are thoroughly compared. 
These contributions are exemplified first on the CIFAR100 dataset, before being scaled up to address the CORE50 continual learning benchmark. 
The final results based on our 3Mb FBNN on CORE50 exhibit at par and better performance than conventional real-valued larger NN models.
\end{abstract}


\section{Introduction}
Binary Neural Networks (BNNs) \cite{qin_binary_2020} have emerged as a promising avenue for enabling Artificial Neural Network (ANN) deployment on ultra-low power devices, \eg, always-on processing for computer vision \cite{verdant_30w5fps_2020, bose_389_2023, garrett_1mw_2023, verdant_450w50fps_2024, zhang_9151220_2023}. 
BNN hardware implementations are highly memory- and energy-efficient with respect to conventional hardware realizations (e.g., floating point or 8-bit integer). 
Furthermore, BNNs are also compatible with a wide variety of emerging hardware technologies, from custom compute-in-memory to emerging optical processing \cite{cho_xnor-vsh_2023, cai_28nm_2023, alhaj_ali_mol-based_2022, shahroodi_lightspeed_2023, sunny_robin_2021}. Addressing both memory and computational constraints makes BNNs ideal for edge-based inference tasks. 

Recently, significant efforts have been made to enhance BNN classification performance through improvements in model structure and training protocols \cite{yuan_comprehensive_2023, sayed_systematic_2023}. These advancements have been built upon the assumption that the training data is identically distributed to the data encountered during inference. 
This assumption falls short in dynamic scenarios, in which the concepts (\eg, classes) targeted for inference (\eg, classification) can evolve or even novel concepts may arise.
Such systems necessitate continual learning~\cite{kumar_sah_continual_2022, machireddy_continual_2022}, \ie, the capability to adapt the model parameters to new concepts without losing the knowledge of old concepts~\cite{kirkpatrick_overcoming_2017}.

The Class Incremental Learning (CIL) methods~\cite{zhou_class-incremental_2024}\cite{icarl} 
address this issue of learning new knowledge while minimizing forgetting in classification problems. 
A relatively simple and effective method for CIL is Experience Replay (ER) that stores a subset of past experiences in a buffer. Whenever the data of a new class (or set of classes, \ie, task) is available, ER retrains the network on a mix of stored data, interleaved with the new-coming data. ER allows the model to learn from both new and old data, thus mitigating forgetting.
Recently, the literature starts to address CIL on BNNs~\cite{onDeviceBNN,vorabbi_enabling_2024}. 
Nevertheless, to achieve sufficient performance, these approaches still require several real-valued network layers. 

To the best of our knowledge, this paper presents the first in-depth study for CIL in Fully Binarized Neural Networks (FBNNs). The main advantage of FBNNs is that their hardware mapping is very efficient. All computations (binarized scalar products with binary inputs) can be performed by highly compact hardware structures, combining point-wise multiplication via logical gates and accumulation using pop-counters~\cite{nguyen_billnet_2022,ma2024b}. 
To this end, we made four key contributions, each exemplified in part on the CIFAR100 dataset. Finally, the overall approach is validated on the CORE50 dataset~\cite{core50}.

Firstly, this paper revisits existing work in~\cite{basso2024class}, thoroughly details the salient points to guide FBNN design for CIL, and extend previous results with new metrics and a comparison with state-of-the-art. 
Most BNNs from the literature are actually not fully binary, with their first and last layers in richer, real-value or integer, representations~\cite{Bannink2020LarqCE}. 
Furthermore, to ease convergence, existing BNNs may include Batch Normalization (BN) layers, typically in floating point.
To avoid channel-wise BN, we propose to train the network in a single-stage, relying on layer-wise scaling factors only as normalization. The values of the scaling factors are set not from the BN layer parameters, but utilising the dimensions of network topology itself. 
Furthermore, for the network's bottleneck we propose a trainable global average pooling to reduce dimensionality of the latent space, making the network robust to spatial translations of the input, despite the low expressivity of binary values.
Last, to avoid real-valued input data, we transform the input image pixels into a luminance/chrominance domain, which is further converted into a thermometer encoding~\cite{thermoCoding}.

Our FBNN is compared with the two closest NN designs, without BN~\cite{Chen_2021_CVPR,ma2024b} or claiming only binary arithmetic~\cite{ma2024b}.
The performance of FBNNs of three model sizes (parameters' memory footprint) is reported: a 3 Mega bit (Mb), a 9Mb, and a 24Mb. Experiments indicate that our network reaches state-of-the-art performance in an offline setting, \ie, when all the dataset is available.
Furthermore, for the CIL experiments, the contributions in this paper mainly focus on our tiny \textit{3Mb-BNN} model. We implement two CIL types of ER, Native replay (storing raw inputs in the buffer) and Latent replay (storing latent features). 

Secondly, we address the main CIL trade-off, namely adaptability vs retention, i.e., performance on past tasks vs. on the current task. 
Loss balancing, a possible strategy to control adaptability and retention, is less addressed in the CIL literature.
Along with loss balancing, here we explore three loss functions suitable to BNNs, namely, the conventional Categorical Cross-Entropy (CCE), the squared-hinge, potentially better for BNNs~\cite{9687541}, and a focal loss. We find that a CCE loss suits well CIL in FBNN, for both Latent and Native replay. Balancing the loss significantly enhances the final performance of all tasks, while slightly decreasing adaptation. 

The proposed FBNN is split in a Feature Extractor (FE) part and a classifier part, as any conventional NN in image classification problems.
The common practice for learning feature representation is to perform supervised learning on a pre-training dataset~\cite{vorabbi_enabling_2024}.
Extracted features can be discriminative for pre-training classes, however, there is no guarantee that they are also discriminative for later tasks. 
It becomes an issue for Latent replay, in which the feature extractor is not updated.

Thirdly, the supervised pre-training approach is challenged, 
proposing a semi-supervised alternative to learn more transferable features in FBNNs. 
Supervised training is combined with Barlow Twin (BT)~\cite{zbontar2021barlow} and custom activation regularization that forces features to be more distributed. As a results, the final test accuracy in increased by 1.17pts in CIL for the CIFAR100.

Fourthly, the FBNN exploration covers both Native replay and Latent replay. The performance of Native replay and Latent replay are compared for a given memory buffer footprint.
For a small memory buffer ($<$25Mb), Latent replay yields higher final accuracy than Native
replay. In this buffer size range, Latent replay offers a better retention but lower adaptation than Native.
For larger buffer sizes, Native is better.

Finally, all the proposed methods are put together and experimented on the CORE50 dataset~\cite{core50}. which serves to benchmark IL strategies in real-life scenarios.
To fit this dataset, we design a deeper architecture, \textit{Res-BNN}, following the principles above, but still having only a 3Mb memory footprint.
Experiments on CORE50 show that our \textit{3Mb-Res-BNN} with Latent replay achieves performance comparable to ER methods using a 352Mb ResNet18~\cite{pham2021dualnet}. More impressively, Native replay surpasses other replay methods, with a model memory footprint that is 119$\times$ smaller and a buffer size 1.8$\times$ smaller, while relying solely on binary scalar products. 

The outline of this paper is as follows.
Section~\ref{sec:rw} discusses the related work for both BNN and CIL. Section~\ref{sec:bg} states the addressed problem and introduces the necessary background and notations. Section~\ref{sec:metrics} defines the evaluation metrics and methodology. 
Section~\ref{sec:NND} introduces the FBNN design guidelines, followed by Section~\ref{sec:lb}, which discusses loss balancing. 
Section~\ref{sec:pt} proposes a semi-supervised pre-training of the FE for CIL in FBNN.
Next, Section~\ref{sec:es} compares Native and Latent replay at iso-memory.
Section~\ref{sec:val} presents the evaluation of the methods on the CORE50 dataset.

\section{BNN and CIL related work}\label{sec:rw}

\subsection{Binary Neural Networks}

Binarization is an extreme case of quantization of NNs, where weights and activations are constraint to 1-bit representation, taking only two values, \eg, $\{+1,1\}$. 
The standard backpropagation training was made possible for BNNs by the introduction of the Straight-Through-Estimator (STE)~\cite{Bengio2013EstimatingOP}\cite{hubara2016binarized}. 
Indeed, to train quantized NNs, Quantization Aware Training (QAT)~\cite{thiruvathukal_survey_2022} keeps a proxy, full-precision version of the training weights, which are quantized at the forward pass. Whereas BNNs achieved significantly close performance to full precision network on simple datasets, the accuracy gap remains large on challenging ones like ImageNet. 

Closing this gap is an ongoing research field. Existing solutions can be categorized as: quantization error mitigation \cite{rastegari2016xnor}, loss function improvement~\cite{hou2016loss}, and reducing gradient error~\cite{qin_binary_2020}.
In addition, BNN model enhancements have been proposed more recently with the introduction of skip connections~\cite{Liu2018BiRealNB}, factorized residual block~\cite{Bannink2020LarqCE} or MUX-based gated residuals~\cite{nguyen_histogram-equalized_2022}, this with extensions to RNN~\cite{nguyen_billnet_2022}. These efforts prioritize classification performance as main figure of merit and while improved accuracy is attained, it often comes at the expense of introducing mixed-precision operations.

Recent approaches further introduce FBNNs, as really hardware-friendly BNNs. Indeed, FBNNs can have a highly efficient implementation with binary-only arithmetic, however they preclude all computation that inherently utilizes real-valued numbers, such as BN. 
BN-free approaches~\cite{Chen_2021_CVPR} train BNN by adjusting adaptive gradient clipping and weight scaling. \textit{AB-BNN}\cite{ma2024b} improves the offline performance of BN-free networks with a mask layer and adapted ReLU. Furthermore, \cite{tang2017train} shows that binarizing the last layer can be done without significant accuracy drop. 

This paper extends FBNNs with CIL capabilities.
Likewise previous approaches, our work demonstrates that BN can be efficiently replaced by scaling factors, without any significant performance loss. 
This paper extends \cite{basso2024class} with a study of loss balancing and semi-supervised feature extractor pre-training.

\subsection{Incremental Learning}

Incremental Learning, often also referred to as continual learning, is a paradigm where a network has to learn continually on ever-changing data distribution. An instance of this problem is when a learner is in charge of comprehending a series of tasks only having access to current task data~\cite{icarl}. 

Several proprieties are desired for such a system: the \textbf{adaptation} on current data, the \textbf{retention} on past distributions, and, to capitalize on past and current knowledge (forward-transfer and backward-transfer~\cite{GEM}). To avoid an unacceptable model size increase, models should grow significantly sub-linearly with respect to the number of tasks.

For classification problems, \cite{vandeven2019scenarioscontinuallearning} defines three scenarios to deal with various data drifts. Class-Incremental Learning (CIL) - addressed in this paper - deals with the occurrence of new concepts (\ie, classes),  Domain-Incremental Learning (DIL) with drift in the input data (given a constant number of classes), and Task-Incremental Learning (TIL) is a simplification of CIL where the model knows the task-label.
 
 The solutions to incremental learning problems can be broadly divided into three major categories~\cite{zhou_class-incremental_2024}: regularization-, architecture-, and rehearsal-based method, 
 even thought the core principle remains the same: regularize the current training to prevent forgetting of past knowledge. 
 
 Regularization-based methods propose to identify important weights relatively to a prior task and prevent their update by adding penalty terms to the loss~\cite{kirkpatrick2017overcoming,pmlr-v70-zenke17a}. Various ``importance'' metrics may be utilized. Even thought such methods only need a snap-shot of weights from the previous iteration, in practice they do not scale up to large networks. 
 
 Architecture-based methods mitigate catastrophic forgetting by constructing task-specific modules inside the network. It is done via parameter isolation where certain part of the network are frozen~\cite{rusu2016progressive} and growing architecture where additional neurons are added to promote adaptation~\cite{yoon2017lifelong}. Such approaches typically require a task-label during prediction. 
 
Rehearsal-based methods mitigate catastrophic forgetting by storing past samples, ``replaid'' during training. They can be divided into: Examplars-Rehearsal which stores a subset of past examples in a memory buffer~\cite{icarl}, and Pseudo-Rehearsal, which uses generative models to synthesize data from past classes in the input~\cite{shin2017continual} or feature space~\cite{dreamnet}.
While Pseudo-Rehearsal reduces memory requirements, it introduces significant challenges for BNNs, especially in generating accurate synthetic binary data. In contrast, Examplars-Rehearsal methods are straightforward to implement, scale well while ensuring stable performance across tasks~\cite{zhou_class-incremental_2024}, making it an ideal framework for studying FBNNs in a CIL setting. 

The usage of Exemplars-Rehearsal for CIL was first proposed on deep NNs in Incremental Classifier and Representation Learning (iCaRL), which uses herding for selecting exemplars, features knowledge distillation and a nearest-mean classifier. EEiL \cite{Castro_2018_ECCV} brings further improvements by utilizing a fully-connected classifier and a balanced fine-tuning stage to correct the bias due to training dataset imbalance. After that, investigations have followed on buffer memory management~\cite{TinyER}, regularization on gradient~\cite{GEM}, knowledge distillation (KD) on logits~\cite{buzzega2020dark}, and KD-induced-bias correction~\cite{Wu2019LargeSI}.

Moreover, there is a renew interest for the Experience Replay (ER) proposed in~\cite{Ratcliff1990}, which consists in interleaving past examples with the current training batch without additional sophisticated techniques or training protocol. Following~\cite{Buzzega2020RethinkingER}, we adopt ER as baseline methods for our simulations. 

A promising alternative to ER is Latent replay, where latent features are stored instead a raw inputs~\cite{memEfficient,latentReplay}. In addition to memory overhead reduction, it decreases the training's computational costs, paving the way to on-device model (re)training~\cite{latentReplay_tinyMl}. Latent replay becomes especially compelling with BNNs, as the latent features are binarized, offering significant storage savings compared to real-values. 

\subsection{CIL relying on BNN}

Lately the literature starts to address CIL on BNNs~\cite{onDeviceBNN},\cite{vorabbi_enabling_2024},\cite{laborieux_synaptic_2021}. Synaptic metaplasticity~\cite{laborieux_synaptic_2021} address catastrophic forgetting in BNN by introducing a metaplasticity-inspired mechanism that modulates weights updates, reducing the likelihood of changes in binary weights that are important for past tasks. This approach lacks scalability, similarly to regularization-based approaches. 

More recently, Vorrabi \textit{et al.} propose to adapt Latent replay with a fixed BNN feature extractor and a quantized classifier with CWR* regularization to mitigate forgetting~\cite{onDeviceBNN}. The work is further extended with more learnable binarized layers, yielding better incremental performance~\cite{vorabbi_enabling_2024}. Nevertheless, this model requires real-valued first, last, and BNs. 

Our work proposes to go a step further, addressing CIL in FBNNs, scaled up to the state-of-the-art~\textit{CORE50} dataset. 
In this paper, we discuss the FBNN design and its training procedure, introduce new metrics to analyze the CIL performance, explore loss balancing, detail a method to pre-train the feature extractor for transferable representations, and, finally, thoroughly compare Latent and Native replay.

\section{BNN-CIL problem formulation}\label{sec:bg}

\subsection{CIL problem formulation and notations}

In class-incremental learning, a model $f(\theta)$, with parameters $\theta$ sequentially learns $T$ classification tasks of training set $D^0$,...,$D^t$,...,$D^{T-1}$ respectively. At task~$t$, starting with weights converged on the previous task, $f$ is optimized on current task $D^t = \{ (x_i^t,y_i^t)\}_{i}$ composed of data-labels pairs drawn from a distribution $\mathcal{T}^t$. An additional dataset $D^{PT}$ is available to pre-train the network. Each task has its own unique set of classes $y_{i}^t \in \mathcal{Y}_{t}$ in the considered CIL scenario. 
For clarity, we denote as offline training the conventional training in which all the data is available at once, when the training is performed.
In class-incremental learning, the retraining at each individual task can be considered as offline training on the current task data. For this dataset the samples can be seen several times and there is no epoch nor batch size restrictions.

Furthermore, we consider that task boundaries are given, meaning there is no need to detect new tasks. 
The objective of the model at task $t$ is to minimize the expected cost on all encountered distributions ${\mathcal{T}^i}_{0\le i \le t}$ as described in (\ref{eq:bg_reg}). However, past distributions are no longer available, we only have access to $D^t$. Even if the goal is respected at task $t-1$, after training on $D^t$ the expected cost on past classes is low.

The CIL problem can be formulated as a regularization problem, as follows: how to to regularize training in order to prevent an increase in the loss on past classes without compromising convergence on the current one. 
The first term in (\ref{eq:bg_reg}) is the conventional classification loss on the available data of the current task.
The second term in (\ref{eq:bg_reg}) is the regularization term that prevents catastrophic forgetting. 
As the previous data in is not available anymore, this term is approximated, by $\mathcal{L}_{ret}$, namely the retention loss. 

\begin{equation}\label{eq:bg_reg}
    \mathcal{L}_{CIL} = \underbrace{\underset{(x,y)\sim\mathcal{T}^t}{\mathbb{E}}[\mathcal{L}(f(x,\theta),y)]}_{\text{available data distribution}} + \underbrace{\sum_{i=0}^{t-1}  \underset{(x,y)\sim\mathcal{T}^i}{\mathbb{E}}[\mathcal{L}(f(x,\theta),y)]}_{\text{approximated by } \mathcal{L}_{ret}}
\end{equation}

\subsection{Experience Replay (ER)}

ER tackles the regularization problem in (\ref{eq:bg_reg}) by keeping a subset of training samples into a bounded memory buffer, referred to as $M$. At each training session, $M$ is concatenated with $D^t$ to form the training set, and hence $M$ introduces an intrinsic regularization term, as depicted in (\ref{eq:bg_er}). 

\begin{equation}\label{eq:bg_er}
    \mathcal{L}_{CIL} = \underset{(x,y)\sim\mathcal{T}^t}{\mathbb{E}}[\mathcal{L}(f(x,\theta),y)] + \underbrace{\underset{(x,y)\sim M}{\mathbb{E}}[\mathcal{L}(f(x,\theta),y)]}_{\mathcal{L}_{ret}}
\end{equation}

As such, the CIL system can be described as a \textit{learner} (the BNN) and a \textit{memorizer} (memory buffer).
There are four aspects that define an ER strategy: how to update the memorizer, what information to store, for the labels as well as for the data samples, and, finally, how to regularize the loss. 

\subsubsection{Memorizer update rule} $M$ stores representative samples from each past tasks. In real use cases, the memory budget is bounded. We thus denote as $\mathrm{m_s}$ the buffer capacity, in number of samples, and $\mathrm{m_b}$ the buffer size, in bits. 
As in the balanced reservoir sampling strategy~\cite{chaudhry2019TinyER}, 
we assume that: (1) $M$ should be fully filled to ensure maximum sample diversity within the memory budget, and (2) every class must be equally weighted.
Within these assumptions, the memorizer randomly chooses: (1) which samples of the current task to store, and (2) which samples of the buffer to replace, when new samples are stored. 

\subsubsection{Raw input vs. latent features storing} Two ER approaches are possible: storing native inputs, \ie, \textbf{Native replay}, or storing latent features, \ie, \textbf{Latent replay}. 
A FBNN, $f(\theta,.) = G(F(\bullet))$ can be divided into two components: a Feature Extractor (FE), $F$ parameterized by $\theta_F$, and a classifier (Clf), $G$ parameterized by $\theta_G$. The FE corresponds to the early layers and have been shown to be transferable across classification tasks, if the initial training dataset shows enough diversity~\cite{Belouadah2018DeeSILDI}. Consequently, if $F$ performs dimensionality reduction, storing compressed latent representations in the memory buffer instead of raw inputs can be advantageous. Given a fixed $\mathrm{m_b}$, in this configuration, more samples are stored, making $M$ more diverse but lowering model expressiveness as $F$ cannot be easily updated~\cite{latentReplay}. The trade-off between diversity and expressiveness is discussed in Section~\ref{sec:es}. 

\subsubsection{Labels vs. logits storing} The training set consists of sample-labels pairs. Nevertheless, other input-output mapping can be stored in $M$. The outputs can be stored as logits or as labels. 
Typically, the logits is the vector of raw, non-normalized, predictions that the classification model produces (inference outputs).
Logits are more informative as they capture classes probability relatively to each other~\cite{Mittal_2021_CVPR}. However, storing logits in CIL requires a bias correction mechanism~\cite{Wu2019LargeSI} and it is not clear if they truly enhance performance~\cite{BELOUADAH202138}. 
Furthermore, storing logits demands more memory than storing labels only, as the logits of each class is an vector of size equal to the number of classes. 
Everything considered, we choose to store labels as output for all our experiments.

\subsubsection{Loss regularization} Several ways to leverage $M$ to regularize the training exist: sample interleaving~\cite{chaudhry2019TinyER}, orthogonal gradient projection~\cite{GEM}. 
For the sake of simplicity and computational efficiency, we choose the sample interleaving approach where batches are formed by randomly drawing samples in $M \cup D^t$. However, current classes are more represented in $D^t$ than past classes in $M$, leading to imbalance. Section \ref{sec:lb} thoroughly discusses loss balancing.

\subsection{The BNN training flow}

A binary network, as any quantized network, requires a special procedure for training, such as Quantization Aware Training (QAT). 
In QAT, the proxy full-precision weights, $\theta^r$, are kept and updated during the training, because binary elements cannot accumulate updates. 
The forward and backward passes are similar to the ones in conventional training with back-propagation, with few differences, as follows.

Typically the forward pass is performed on the quantized weights. To do so, every full-precision value,  $x^b$, is binarized with the \textsc{sign} function (\ref{eq:sign}), resulting in $x^b$. 

\begin{equation}\label{eq:sign}
      x^b = \textsc{sign}\left( x^r \right) = 
            \begin{cases}
                +1,& \text{if } x^r \geq 0\\
                -1,& \text{otherwise}
            \end{cases}
\end{equation}

As \textsc{sign} is not differentiable, in the backward pass its derivative is approximated by a clipped identity (STE)~\cite{Bengio2013EstimatingOP}: 

\begin{equation}\label{eq:clip_id}
    \frac{\partial \mathcal{L}}{\partial \theta^r} \simeq 
    \frac{\partial \mathcal{L}}{\partial \theta^b} \mathds{1}_{[-1;+1]}
\end{equation}

Special attention should be given to BN~\cite{pmlr-v37-ioffe15}, an efficient method to ease model convergence by properly scaling the values in forward and backward passes. 
BN yet relies on full-precision operations that cannot be easily binarized. 

A typical QAT multi-stage training~\cite{nguyen_histogram-equalized_2022},~\cite{ma2024b} is as follows: (1) train with quantized weights, full-precision activations, and BN, (2) train with quantized weights, quantized activations, and BN, and (3) train with quantized weights, quantized activation and scaling factors initialized from BN parameters. 

On the last stage, the scaling factors can be computed as an average or median of the estimated moving averages and internal gain parameters. 
Offsets and biases are discarded as adding values to a binary +1/-1 may flip the sign and introduce large errors. 
The role of the scaling factors is to help to converge the training when the forward pass is performed only with binary values. The scaling factors are hence removed for inference, since it has no impact on the $\textsc{sign}$ function. 

\section{Metrics and evaluation methodology}\label{sec:metrics}

In the CIL community, the goal is a high final accuracy, after a given number of learned tasks. Previous work investigates the final accuracy along two directions, adaptation and retention, as defined below. We argue that other metrics allow to better understand the details of a CIL system, as presented in the core of this section. Finally, our experimental setup with respect to CIL and FPNN is extensively detailed. 
\subsection{Conventional metrics}

The three important properties of CIL are the ability to adapt to new classes (adaptation), to remember past classes (retention), and to generalize well on all seen classes (global accuracy). These can be evaluated by the accuracy of the corresponding subset of classes, on the test set. \textbf{Adaptation} is measured as the accuracy on the current, \ie, new task $D^t$, and denoted as $\mathrm{a_{new}}$. \textbf{Retention} is measured as the accuracy on the first task $D^0$, and denoted as $\mathrm{a_{old}}$. We compute it on $D^0$ rather than on all past classes $\mathcal{Y}_{i < t}$ to give a indication on the forgetting of the oldest task. 
\textbf{Global accuracy} $\mathrm{a_{seen}}$ is the accuracy on all seen classes $\mathcal{Y}_{i \leq t}$. For the sake of clarity, we denote as $\mathrm{a_{final}}$ the global accuracy after all tasks are learned. 

\subsection{Beyond conventional metrics}

Our learning system comprises a compact FBNN and a bounded memory buffer, both introducing challenges in training. Table~\ref{tab:metrics} summarizes relevant metrics notations. 
The training accuracy is often omitted in the literature, whereas it offers significant insights on the model expressiveness, providing useful information on overfitting issues. 
We report the accuracy figures for the train and test cases, when necessary.
As the FBNNs studied are compact, a plateau of the training performance may indicate that the test performance is capped by the size of the network, whereas a large training performance may suggest that further regularization could increase test performance. 

Furthermore, the buffer size should be low (\eg, $\le 100$Mb), implying a high risk of overfitting. 
To quantify this, we introduce two metrics: $\mathrm{a^{train}_{buffer}}$, the accuracy computed on the buffer samples in the train set, $M$, and $\mathrm{a^{test}_{buffer}}$, the accuracy computed on the test set of past classes $\mathcal{Y}_{i < t}$, at task $t$.

Finally, the network may be initially pre-trained on a dataset $D^{PT}$ and then incrementally retrained. To have an indication of the performance at the start of the incremental retraining we also introduce the accuracy at the pre-training task $a_{PT}$.

\subsection{Dispersion metric}

The CIL problem can be framed as a detection problem, instead of a classification problem, where the model's goal is to detect new classes, as well as past classes. The accuracy on a subset of classes only gives the average of classes recall. To provide more insight about each class detection, we propose a dispersion metric, denoted as $\mathrm{d}$, defined as the standard deviation on class recalls. If the model must detect every class equally well, the ideal case is the one where the dispersion is minimal. Combined with the accuracy, dispersion also gives insights on the lower-bound detection performance. Lower accuracy with smaller dispersion can be preferable to higher accuracy with bigger dispersion, if the minimum detection performance is higher. As with accuracy, dispersion can be estimated on a subset of the data, as visible in Table~\ref{tab:metrics}.

\begin{table}[]
 \caption{Notations for the accuracy, $\mathrm{a}$, and dispersion, $\mathrm{d}$, metrics. The evaluation is done on test set when the superscript is omitted.}
\centering
\label{tab:metrics}
\begin{tabular}{|c|c|l|}
\hline
{subscript}   & $\mathrm{PT}$     & pre-training task $D^{PT}$   \\ \cline{2-3} 
                              & $\mathrm{old}$    & first task $D^0$             \\ \cline{2-3} 
                              & $\mathrm{new}$    & current task $D^t$           \\ \cline{2-3} 
                              & $\mathrm{seen}$   & all encountered tasks        \\ \cline{2-3} 
                              & $\mathrm{final}$  & all tasks                    \\ \cline{2-3} 
                              & $\mathrm{buffer}$ & samples in buffer $M$        \\ \hline
{superscript} & $\mathrm{train}$  & samples seen during training \\ \cline{2-3} 
                              & $\mathrm{test}$   & samples in the test set      \\ \hline
\end{tabular}
\end{table}

\subsection{Evaluation Methodology}\label{sec:eval_methodo}

\subsubsection{Datasets} 

In the rest of this paper, the FBNN-CIL is exemplified on \textit{CIFAR100}~\cite{krizhevsky2009learning}. This is composed of labeled images drawn from 100 classes, each image having $32\times32$ pixels, with 3 color channels, each encoded on 8-bit integers. Each class is composed of 500 training images and 100 test images. We keep $10\%$ of training images for validation.  CIFAR50+5X10 denotes the following scenario: the 50 first classes are used for pre-training $\mathcal{Y}_{PT} = (0,...,49)$ and the 50 remaining classes are split into 5 tasks of 10 classes. When possible, the training is regularized using image data augmentation (\ie, random flip, random translation, and random contrast). 
Section \ref{sec:val} finally reports an in-depth evaluation of the proposed approach on \textit{CORE50}~\cite{core50} with a pre-training on \textit{STL10}~\cite{coates2011analysis} for a dedicated FBNN (\textit{3Mb-Res-BNN}).

\subsubsection{Hyperparameters} 

 We employ the Adam optimizer, with an adaptive Reduce-On-Plateau learning rate scheduler \cite{chollet2015keras}, and an early-stop \cite{prechelt2002early} for each retraining stage. 
 \ref{sec:lb} discusses in detail the type of loss, considering Categorical-Cross-Entropy (CCE) as the baseline. Optimization is performed with batch size of 64 and an initial learning rate of $10^{-4}$. A small batch size combined to a low initial learning rate were used, as we observed that higher ones may cause training unreliability.   

\section{Guidelines for FBNN design in CIL}\label{sec:NND}

\begin{figure}[!h]
\centering
\includegraphics[trim=0 0 0 0,clip, scale = 0.5]{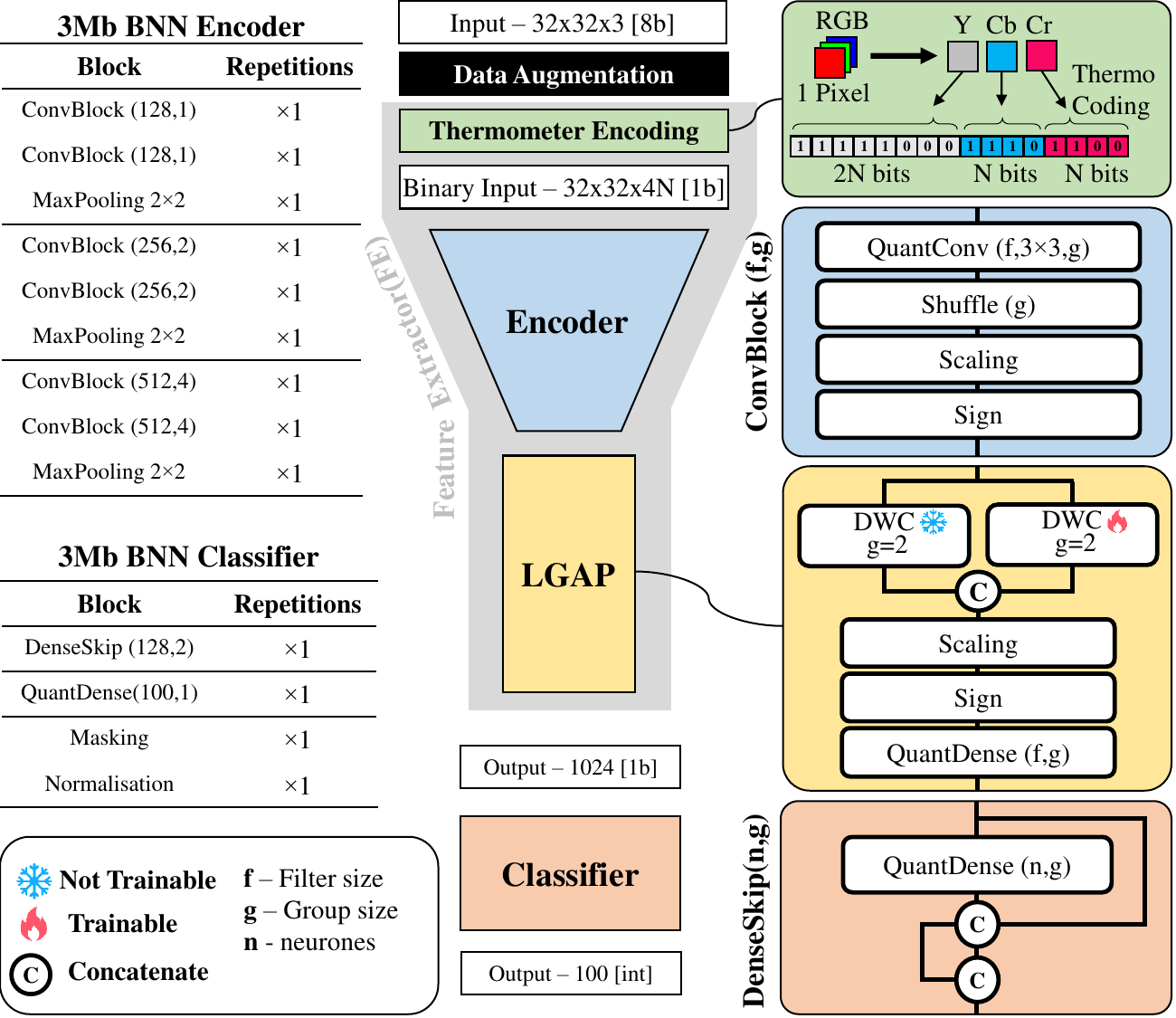}
\caption{Baseline model structure. \textit{3Mb-BNN} version reported here.}
\label{baseline:modelstructure}
\end{figure}

Despite their efficient execution, BNNs are challenging to design and train~\cite{QIN2020107281}.
The literature proposes various methods for sizing, designing and optimising BNNs~\cite{Bannink2020LarqCE}~\cite{Liu2018BiRealNB}.
This section revisits and extends related work~\cite{basso2024class}, by stating 5 key design practices to accommodate CIL with ER applied to FBNNs. 
For the sake of simplicity and repeatability, a basic (but efficient) VGG-like model head \textit{3Mb-BNN} (Fig.~\ref{baseline:modelstructure}) has been first considered. The model head is composed of 3 convolution blocks, each composed of a stack of two g-grouped convolutions~\cite{Shufflenet} followed by a 2$\times$2 Max Pooling. The number of filters and groups (f,g) are doubled at each block, \ie, (128,1), (256,2) and (512,4). This FE head is followed by a dedicated custom bottleneck and classifier which are described in the following.

\subsection{Key BNN enablers for CIL}\label{sec:key_cil}

\subsubsection{Scaling factors as normalization} 

In the offline setting, layer-wise scaling factors can efficiently replace channel-wise BN without performance drop by employing multi-stage QAT. In incremental learning, multi-stage QAT would imply re-initializing BN parameters at the beginning of each retraining. However, BN has been shown to exacerbate catastrophic forgetting by having its parameters, means and variances, biased towards over-represented new classes in training batch. task~\cite{cha2023rebalancing,lomonaco2020rehearsal}. Thus it is preferable to not derive scaling factors from the BN parameters in a multi-stage QAT during retrainings. 

To circumvent this issue, we propose to train the network in a single-stage, and thus set the scaling factors' values not from the BN layer parameters, but utilising the network topology itself, making them task-agnostic. Inspired by the supposed BN's goal of limiting the covariance shift~\cite{pmlr-v37-ioffe15}, we aim for scaling factors that give unitary activation variance. 

At the initialization, since weights and activations are assumed to be equally distributed between -1 and +1, it means that activation $A^b$ and weights $\theta^b$ follow a Rademacher distribution of parameter $p = 0.5$. The resulting dot product between $\theta^b$ and $A^b$ thus follows a binomial law of mean $0$ and variance $Var = 2^2np(1-p)=n$, where $n$ corresponds to number of input of the dot products, also called $FanIn$. An appropriate scaling factor $S$ could therefore be defined depending on $FanIn$ and a certain scalar $K$: 

\begin{equation}\label{eq:scaling-factor}
      S = \frac{K}{\sqrt{FanIn}}
\end{equation}

We conducted experiments to determine the best value for $K$ for the \textit{3Mb-BNN} model, so that maximum train accuracy is achieved. Maximum train accuracy is desired to leave room to the network to optimize further by regularization such that good test performance can be also be reached.
Table \ref{tab:1-normalisation} presents the train accuracy for a case with and without BN and for three $K$ values. 
The unitary variance ($K=1$) brings comparable results to the use of BN, in a single-stage training.

\begin{table}[!h]
\centering
\caption{Influence of variance scaling in normalization layer}
\label{tab:1-normalisation}
\begin{tabular}{|c|c|c|c|c|}
\hline
&BN & $K$=0.3 & $K$=1 & $K$=3 \\ \hline
Train accuracy &78.7\%  &  74.7\%     &  78.5\%   &  75.5\%   \\ \hline
\end{tabular}
\end{table}

Special attention should be given to the scaling of the last layer.
Experiments indicate that instabilities may occur during training without BN when the number of classes increases. After retraining for several successive tasks, the training loss tends to cap.
The reason behind may be that the network is fully binary, and hence the $\{-1, 1\}$ representation in the last layer lacks dynamic range, which may result poor output gradient scaling.
We find that, what we call an $\alpha$-scaled binarization, \ie, $\alpha \times \textsc{sign}(\dot) $,
at the end of the model is beneficial for convergence, as it scales the output gradient. $\alpha$ is dependent on the number of classes, $N_c$, as follows.

\begin{equation}\label{eq:output_scaled}
    \alpha = \frac{1}{\sqrt{c \times FanIn \times N_c}}
\end{equation}

Similar to the scaling that replaces BN, this scaling is employed during training and removed during inference.
The results in Table \ref{tab:1-normalisation} are on CIFAR100, thus with $N_c=100$, and with the constant $c$ empirically set to $5$.
The single training procedure is applied at each CIL retraining step. Nevertheless, this single training procedure does not preclude a pre-training stage, when it appears crucial. The weights initialization can be done with any conventional method~\cite{glorot2010understanding}, as well as relying on a proxy pre-training.

\subsubsection{Bottleneck design} Including a Global Average Pooling layer (GAP) at the end of the convolutions blocks is a common practice to both reduce the dimensionality of latent representation before the classifier and make the network robust to spatial translation of the input~\cite{Lin2013NetworkIN}. However, with binary features as input, the importance of a given spatial region cannot be encoded thought the activation magnitude. To keep spatial information at the output of the FE we propose to add a path where the pooling weights are learned. The whole projection block is noted LGAP (learnable GAP) and depicted in Fig.~\ref{baseline:modelstructure}. In practice, binary weighting is learned as a Depth Wise Convolution of the spatial size of input filters. Both projections are concatenated, scaled, and binarized. Finally, features are recombined by a dense layer in a $1024$ dimension binary latent space. It results in a compression factor of $24$ compared to the input 8-bit image.

\subsubsection{Input data encoding} A common practice in BNN design is to keep the first layer in full-precision at the cost of its computational overhead. Pixel normalization, centering and standardization are thus used to pre-process the input. We aim for binary-only arithmetic for each layer, comprising the first and last ones. \cite{thermoCoding} proposes to code each $RGB$ channel with a thermometer coding, denoted as $TRGB$, on 85 binary channels after a uniform quantization. This leads to better performance than RGB with pixel normalization. 
Each pixel in an image thus requires $3 \times log_2(\lceil 85 \rceil) = 21$ bits of storage in the case of ER with a buffer. 

In a fully binary implementation, the hardware mapping can be even more eased when the number of bits per pixel are an integer power of 2.
To further optimize the input data encoding, $RGB$ values are first transformed into a $YCbCr$ (YCC) color space and then converted into a thermometer coding, $TYCC$. Since the luminance channel, Y, contains more discriminative information that the chrominance channels, $CbCr$, we allocate 2 times more channels to the Y ($2N$) than to the C channels ($N$). 
Experimentally we determined that $TYCC$ with $N=16$ C channels gives the same performance as $TRGB$ for CIFAR-100 in offline training, and, in addition, it requires 25\% less number of bits per pixel. In the rest of this paper we hence utilize $TYCC$ with $N=16$. 

\subsubsection{Classifier depth} We consider that the classifier begins after the LGAP block. In Latent replay, this defines the frontier between trainable and frozen layers. The question that arises is: how deep the classifier should be?  On the one hand, many approaches choose a single dense layer~\cite{hadsell2020embracing} for a classifier. Hence the FE output must be linearly separable and thus it is easy to analyze. Finally, a single dense layer eases the optimization as a stack of dense layers is harder to optimize. On the other hand, in case of Latent replay, a single layer does not leave a lot of room for adaptation to future tasks. 
We propose to use a multi-layer classifier with skip connections as depicted Fig.~\ref{baseline:modelstructure} with the DenseSkip block with $n$ neurons and $g$ groups. In a full precision dense layer, skip connections provide shortcuts ensuring the classifier will perform at least as well as a linear one. Here, binary weights values do not allow null projections and thus do not allow path gating. To that end, we duplicate the input activations, so that the $0$ value can be obtained by opposite sign weighting on the same activation node. Finally, grouped dense layers (with 2 groups) are introduced to cap the classifier model size.

\subsection{FBNN evaluation}

This section first evaluates the proposed FBNN \textit{3Mb-BNN} in an offline setting to put in perspective the performance in comparison with state-of-the-art BNNs, as well as with their full-precision counterparts.
Second, we define and evaluate two CIL baselines, that represent the lower- and upper-bound performance for any incremental learning strategy.

\subsubsection{Baselines}

In the offline setting, in order to assess the effect of binarization, we evaluate our $\textit{3Mb-BNN}$ against two floating-point variants: (1) a memory-footprint equivalent network, $\textit{FPNN}_{b}$, and (2) a topology equivalent network $\textit{FPNN}_{p}$. Therefore, 
$\textit{FPNN}_{b}$ and $\textit{FPNN}_{p}$ have real-valued weights and activations in a floating-point (\eg, considered coded on 32b).
On the one hand, $\textit{FPNN}_{b}$ keeps the same topology as our $\textit{3Mb-BNN}$ but has $32\times$ less parameters. 
The number of parameters reduction is done by scaling down the number of filters. For a fair comparison, $\textit{FPNN}_{b}$ uses BN and conventional input normalization between $[-1,+1]$ (no thermometer coding). 
On the other hand, $\textit{FPNN}_{p}$ has the same architecture and the same number of parameters as $\textit{3Mb-BNN}$, and hence a $32\times$ larger memory footprint while using BN. This evaluation (Fig.~\ref{baseline:evaluation}) comprises three network sizings: 3~Mb (\textit{Tiny}), 9~Mb (\textit{Medium}), and 24~Mb (\textit{Large}). Fig.~\ref{baseline:modelstructure}  presents $\textit{3Mb-BNN}$, while 9~Mb and 24~Mb FBNNs are variants obtained by multiplying respectively by 2 and 4 the number of filters. 
$\textit{3Mb-BNN}$ is also compared to \textit{BNN-BN}~\cite{Chen_2021_CVPR} and \textit{BNN-AB}~\cite{ma2024b}, two training approaches without BN but with input and output layers using full-precision. \textit{BNN-BN} an \textit{BNN-AB} test accuracy results both rely on using two different backbones: BiResNet18~\cite{Liu2018BiRealNB} (11.7~Mb) and ReActNetA~\cite{liu2020reactnet} (29.3~Mb).

In the CIL setting, we evaluate our model on the CIFAR50+5X10 scenario on four experience replay baselines: \textit{naive}, \textit{naive-reset}, \textit{cumulative} and \textit{cumulative-reset}. 
The \textit{naive} baselines correspond to the case of retraining only with the new task data, without any incremental learning strategy, hence highlighting catastrophic forgetting; they are thus equivalent to experience-replay with an empty buffer.
The \textit{naive} baselines offer a lower bound on the performance of any incremental learning strategy. 
The \textit{cumulative} baselines correspond to the case of an infinite memory buffer, where all past labelled data are stored, hence indicating the upper-bound performance. 
The \textit{-reset} suffix indicates that weights are randomly initialized at the beginning of each retraining. The \textit{reset} baselines assess the influence of parameter initialization.

\subsubsection{Offline comparison: binary vs. 32-bits floating-point} 
\begin{figure}[h]
\centering
\includegraphics[trim=0 90
0 0,clip, scale = 0.4]{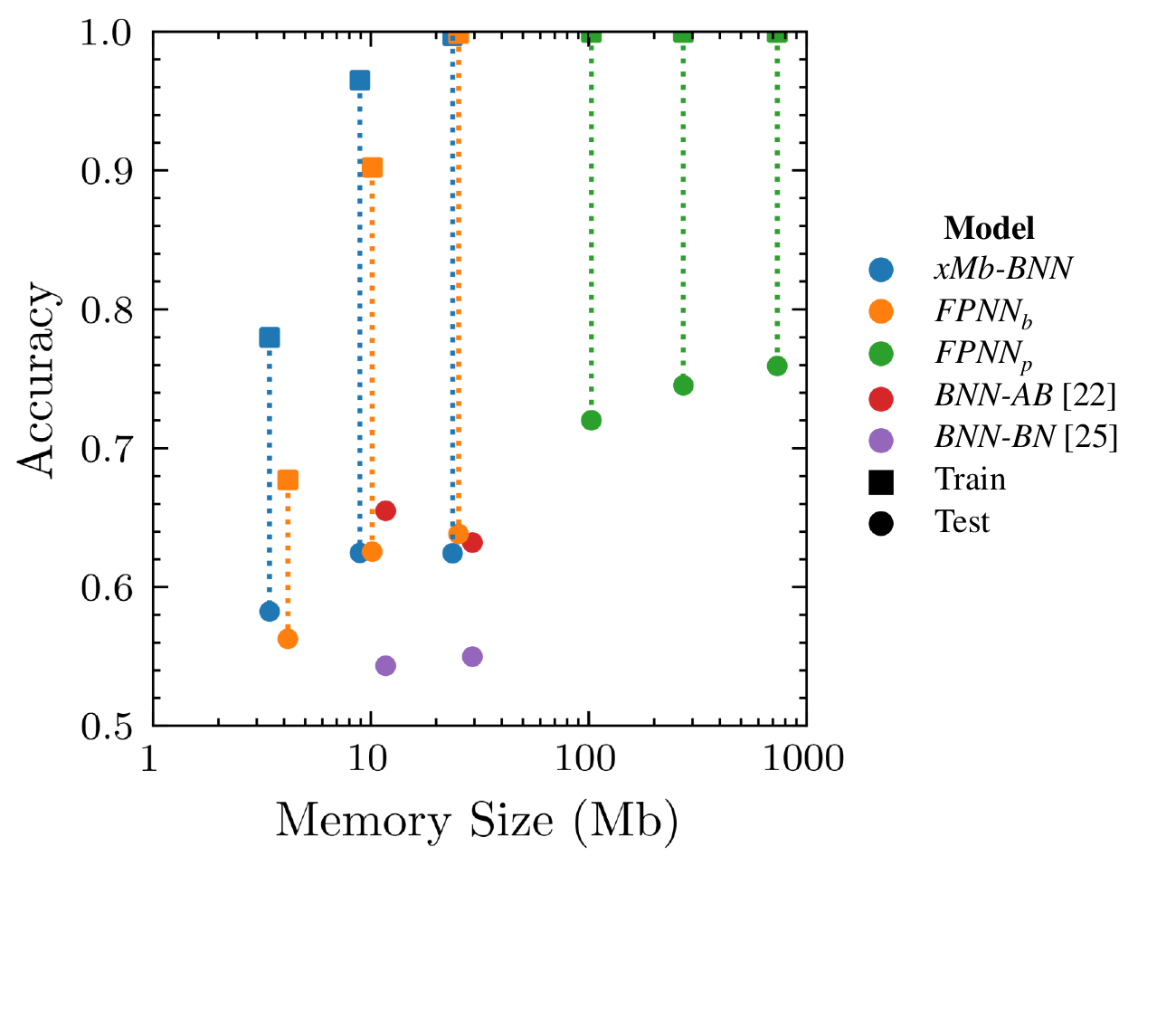}
\caption{Offline accuracy for different configurations of our architecture. Test accuracy is reported with circle and train accuracy with squares.}
\label{baseline:evaluation}
\end{figure}

Fig.~\ref{baseline:evaluation} reports offline accuracy as a function of the model size (Mb), indicating that $\textit{3Mb-BNN}$ provides better performance than \textit{$FPNN_b$} on train set and test set for the 3~Mb and 9~Mb configurations. 
We also note that the 24~Mb variant does not improve test accuracy compared to the 9~Mb, but it still improves the training accuracy. This plateau where increasing the model size above 10-Mb does not lead to better test accuracy, is common~\cite{Bannink2020LarqCE}. Large BNNs seem more difficult to regularize.   
$\textit{FPNN}_{p}$ models achieve higher test accuracy by a bit more than 10\% compared to our FBNNs, while training accuracy remains capped to 100\% (confirming results of~\cite{QIN2020107281}). 

When compared to existing BNNs without BN, our \textit{3Mb-BNN} model shows better performance than~\cite{Chen_2021_CVPR} with more than $3\times$ less parameters, and similar performance to~\cite{ma2024b} in the \textit{Large-BNN} configuration. In conclusion, our FBNN reaches state-of-the-art performance in this particular offline setting.

In all the experiments of this paper we only consider \textit{3Mb-BNN}, unless explicitly stated otherwise.
Moreover, since the training accuracy does not cap to 100\%, the \textit{3Mb-BNN} is the most challenging candidate to study compact FBNN for CIL. It particularly allows to evaluate the forgetting issues in the context of a very limited expressiveness of the model.

\subsubsection{CIL baselines analysis}

\begin{figure}[h]
\centering
\includegraphics{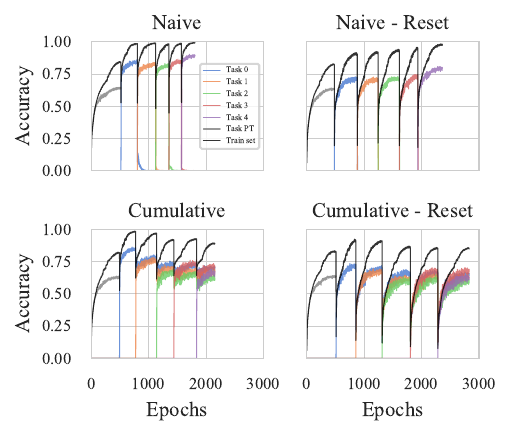}
\caption{Training curves of \textit{3Mb-BNN} on the CIL CIFAR50+5X10, for the 4 baselines strategies. Accuracy on train set is reported on black, pre-training test accuracy in grey, and the test accuracy of each task in a different colors.}
\label{fig:baselines_cil}
\end{figure}

Fig. \ref{fig:baselines_cil} reports the accuracy, for the four baselines, on each task, per epoch, during the whole CIFAR50+5X10 incremental dataset, for \textit{3Mb-BNN}. 
Table \ref{tab:baseline_cil} complements the results with metrics at the end of all 5 retrainings. 
When looking at the validation curve on the current task dataset, we can notice that our training protocol enables a robust adaptation at each task and for each baseline.

As expected, the \emph{naive} baselines retain only the task of the last retraining, whereas in the  \emph{cumulative} baselines all tasks are learned. 
For each task, $t$, the validation accuracy on the current task $D^t$ is higher for \textit{naive} than \textit{cumulative}. For \textit{cumulative}, the training includes every past task's training set, $D_0,...,D_t$, whereas \textit{naive} includes only the current task, $D_t$. Therefore, simultaneous learning of multiple tasks does not contribute to improve learning performance for individual tasks.
Resetting the weights in the \emph{naive} baseline has little influence on the performance; the network forgets all the tasks except the last one regardless on how the weights are initialized.

On the cumulative baselines, weight initialization has a crucial impact. When the weights are initialized with their value from the previous training, the network achieves both higher accuracy on current and old tasks. The same result is observed in Table~\ref{tab:baseline_cil}. $\mathrm{a_{new}}$ and $\mathrm{a_{old}}$ are significantly higher without \textit{reset}. This indicates that \textit{3Mb-BNN} benefits from weight initialization from the previous task, to increase performance on following tasks.
At the end of all tasks, the \textit{cumulative} setting has a $3$pts increase in final accuracy compared to the \textit{cumulative-reset} setting. Moreover, the number of epochs is reduced by $41\%$ for the entire retraining for the 5 tasks when the weights are not reset. 
It indicates our FBNNs capitalize on past knowledge to better learn new tasks. 

\begin{table}[]
\centering
\setlength{\tabcolsep}{2pt}
\caption{Task-level metrics during last retraining (Task 4).}
\label{tab:baseline_cil}

\begin{tabular}{|c|c|c|c|c|c|c|c|}
\hline
baseline         & $\mathrm{a_{old}}$ & $\mathrm{d_{old}}$ & $\mathrm{a_{new}}$ & $\mathrm{d_{new}}$ & $\mathrm{a_{final}}$ & $\mathrm{d_{final}}$ & epochs \\ \hline
naive            & 0.0\%     & 0         & 89.8\%   & 0.06    & 17.9\%     & 0.36      & 236    \\ \hline
naive-reset      & 0.0\%     & 0         & 79.7\%   & 0.07    & 15.9\%     & 0.32      & 428    \\ \hline
cumulative       & 71.0\%    & 0.16      & 67.1\%   & 0.12    & 67.4\%     & 0.15      & 318    \\ \hline
cumulative-reset & 66.0\%    & 0.14      & 62.9\%   & 0.13    & 63.9\%     & 0.14      & 545    \\ \hline
\end{tabular}
\end{table}

\section{Loss balancing}\label{sec:lb}

In general, we can consider that each incremental training session with a replay buffer is a training on an imbalanced dataset containing all seen classes $D^t\cup M$. 
Typically, there are more samples per class in the dataset of the new task $t$, $D^t$, than in the buffer with past task's samples, $M$. The common practice to compensate for an imbalanced dataset is to weight the loss. 
However, CIL introduces two additional biases that can challenge this common practice. The first bias concerns weight initialization at each retraining, which is biased toward past classes. The second bias concerns pre-training of model weights in Latent replay. Indeed, some real-life use-cases may require to retain the classes in the pre-training task. As the feature extractor is trained on $D^{PT}$ then fixed, there is a strong bias toward the retention of the pre-training task. 

As methods to deal with these two biases, in what follows we investigate: (1) the influence of inverse-frequency loss class-balancing, (2) the utilization of a focal term, and (3) the choice of the classification loss. 
These three factors are studied for two use-cases, Retained Pre-training Task (RPT) and Forgotten Pre-training Task (FPT), which, as the name suggests, deal with remembering of not the pre-training task.

\subsection{Losses and loss-balancing strategies}

Inverse class-frequency adjustment is a common strategy to solve the class-balancing problem. The loss terms associated to training samples of class $i\in \left\{1, ..., C\right\}$ are weighted by a factor $w_i$ computed as defined in (\ref{eq:weighti}). We note the class-cardinality and the total number of samples in the re-training dataset, $D^t \cup M$, as $n_i$ and $N$, respectively. $f_i$ additionally denotes the frequency of a class $i$ as in (\ref{eq:fi}). 

We evaluate the inverse class-frequency weighting for three losses: Categorical Cross-Entropy (CCE), Focal Categorical Cross-Entropy (FCCE), and Squared-Hinge (SH).

\begin{align}
    w_i &= C \times \frac{(f_i)^{-1}}{\sum_{j=1}^C (f_j)^{-1}} \label{eq:weighti} \\
    f_i &= \frac{n_i}{N}\label{eq:fi}
\end{align}

The CCE is conventionally defined as in (\ref{eq:cce}) where $B$ is the batch size, $y_{i,c}$ the one-hot encoded label for sample $i$ and class $c$ and $p_{i,c}$ the predicted probability for class $c$ for sample $i$ obtained after applying \textit{softmax} function to logits. 

\begin{equation}\label{eq:cce}
    \mathcal{L}_{CCE} = -\frac{1}{B}\sum_{i=1}^{B}\sum_{c=1}^{C} w_c .y_{i,c}.log(p_{i,c})
\end{equation}

The FCCE loss~\cite{ross2017focal} is a variant of the standard categorical cross-entropy loss that is designed to address class imbalance by down-weighting the loss assigned to well-classified examples, focusing more on harder-to-classify examples. FCCE introduces a tunable focusing parameter $\nu$ that adjusts how much attention is paid to difficult examples. (\ref{eq:focal}) defines FCCE using the same notation as above: 

\begin{equation}\label{eq:focal}
    \mathcal{L}_{FCCE} = -\frac{1}{B}\sum_{i=1}^{B}\sum_{c=1}^{C} w_c (1-p_{i,c})^{\nu}  y_{i,c}.log(p_{i,c})
\end{equation}

The SH loss, designed for Support Vector Machines (SVM), has shown promising results with BNN~\cite{9687541}. Moreover, as many CIL methods use margin loss for classification~\cite{zhou_class-incremental_2024}, we consider it in the loss comparison for BNN-CIL. SH penalizes miss-classifications with a quadratic cost, leading to a smoother gradient and encouraging larger margins between classes for better classification performance. (\ref{eq:sh}) defines SH where the label $y_{i,c}$ is in $\{-1,+1\}$ (1 if the sample belongs to class $c$, -1 otherwise) and $\hat{y}_{i,c}$ is the associated logits.

\begin{equation}\label{eq:sh}
    \mathcal{L}_{SH} =  \frac{1}{B}\sum_{i=1}^{B}\sum_{c=1}^{C} w_c .\left( max\left(0, 1-\hat{y_{i,c}}. y_{i,c}\right) \right)^2
\end{equation}

\begin{figure}[!h]
    \centering
    \subfloat[Native replay - RPT]{\includegraphics[width=0.5\textwidth]{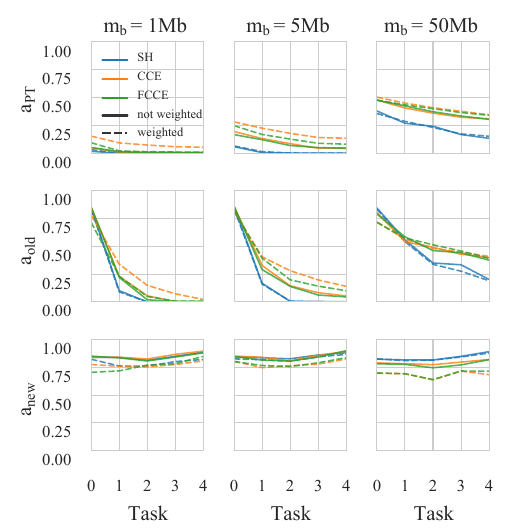}\label{fig:nr_rpt}}
    \hfil
    \subfloat[Latent replay - RPT]{\includegraphics[width=0.5\textwidth]{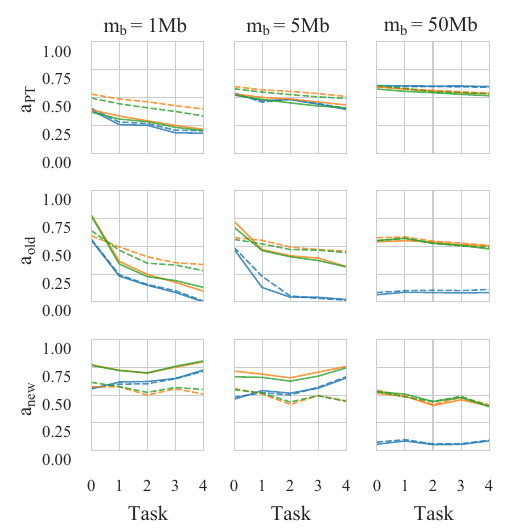}\label{fig:lr_rpt}}
    \\
    \subfloat[Native replay - FPT]{\includegraphics[width=0.5\textwidth]{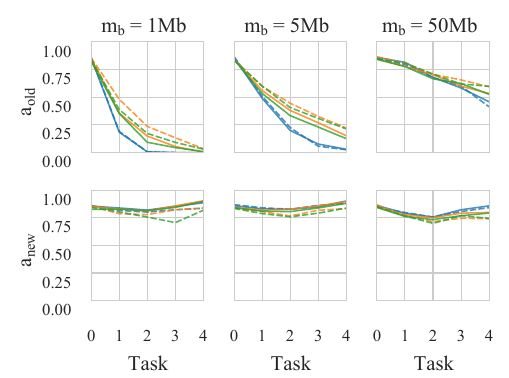}\label{fig:nr_fpt}}
    \hfil
    \subfloat[Latent replay - FPT]{\includegraphics[width=0.5\textwidth]{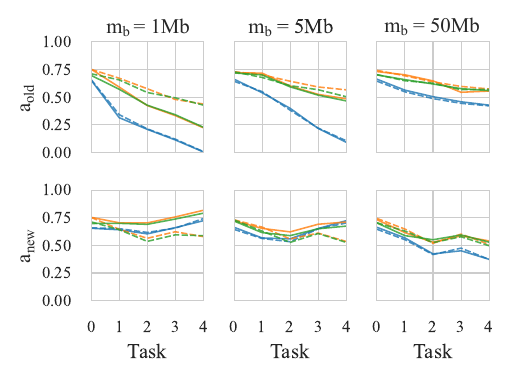}\label{fig:lr_fpt}}
    \caption{Influence of loss and buffer size in RPT and FPT scenarios. Latent replay and Native replay strategies are reported with various losses (colors).}
    \label{loss_balancing:results}
    \label{fig:balancing}
\end{figure}

\subsection{Experimental results}

\textit{3Mb-BNN} is evaluated on CIFAR50+5X10 for both Native and Latent replay. Unlike FPT, for RPT, the classifier is not reset after pre-training and the memory buffer includes samples from $D^{PT}$. As in~\cite{ross2017focal}, we set $\nu = 2$ for FCCE loss.

\subsubsection{The SH loss} With respect to Native replay, Fig. \ref{fig:nr_rpt} and \ref{fig:nr_fpt} indicate that SH leads to better performance on the current task, $\mathrm{a_{new}}$, than other losses at the expense of deteriorating retention, $\mathrm{a_{old}}$, regardless of the buffer size. SH seems to exacerbate plasticity to the point of degrading the final accuracy. Figures also indicate that SH is relatively insensitive to class-weighting, which is probably because, by definition, it attempts to find margins under a one-vs-all classification problem statement. 
In Latent replay, even with a $50$Mb buffer, the SH loss exhibits convergence issues on current tasks while the accuracy on the pre-training stays high. 

\subsubsection{The cross-entropy losses} Let's first analyze the adaptation and retention capacity in Fig.~\ref{fig:balancing}. As expected, inverse frequency balancing improves retention, \ie, $\mathrm{a_{old}}$, for both Native and Latent replay, especially with smaller buffers. By giving higher weight $w_i$ to the under-represented past classes, the model preserves knowledge from earlier tasks, enhancing retention. However, this comes at the cost of reduced adaptation to new classes, \ie, $\mathrm{a_{new}}$, as the increased focus on past data limits the model's ability to adjust to new information. Latent replay is more sensitive to balancing. For instance, we can observe a $41.0\%$ reduction in adaptation on the last task for Latent replay compared to a $7.7\%$ reduction for Native replay with CCE loss in RPT scenario $m_b$=5~Mb. 

Secondly, the final performance is shown in Table~\ref{tab:RPT} and Table~\ref{tab:FPT} for RPT and FPT. For clarity, we report final metrics for Latent 5~Mb and Native 50~Mb, \ie, a similar buffer capacity in terms of number of samples. In both configurations, approximately 50 samples per class are stored. In both configurations, loss balancing leads to improvement of $a_{final}$. Particularly in the RPT scenario, loss balancing increases the final performance by at least $+5$pts, in Latent replay (5Mb) for both losses. Weighting the loss decreases class dispersion by a factor $9.7\%$ in average for FPT scenario and by $18.2\%$ in RPT scenario. Loss balancing is thus effective at both enhancing final accuracy and reducing class dispersion in both scenarios.   

\subsubsection{Loss focalization} The focal loss does not seem to bring any substantial advantage in the final accuracy (Table \ref{tab:FPT} and \ref{tab:RPT}). However, it reduces classification dispersion at the cost of small final accuracy decrease. If the use-case where the CIL is employed requires that each task is classified with (more or less) the same accuracy, one can consider utilizing FCCE.

\subsubsection{Main takeaways} Despite being a good practice in offline learning in BNN, the SH loss is not highly relevant for FBNN-CIL as it decreases the final accuracy. A CCE loss suits well FBNN-CIL. In this case, balancing the loss significantly enhances performance in every configuration while slightly compromising the adaptation. \ie, decreasing the accuracy on the new task. The dispersion in per-class performance can be reduced by adding a focal term.
Latent replay appears more sensitive to the type of loss especially on a small buffer (\eg, 1Mb) than Native replay. 
In the rest of this paper all experiments use the CCE loss with inverse-frequency class-weighting.
In the following sections we focus on FPT, where the pre-training only learns the FE. 

\begin{table}[]
\centering
\caption{Comparison of balancing strategies in the RPT scenario.}
\label{tab:RPT}
\begin{tabular}{|c|c|cc|cc|}
\hline
{strategy}      & {loss} & \multicolumn{2}{c|}{weighted}                   & \multicolumn{2}{c|}{not weighted}              \\ \cline{3-6} 
                &        & \multicolumn{1}{c|}{$\mathrm{a_{final}}$} & $\mathrm{d_{final}}$  & \multicolumn{1}{c|}{$\mathrm{a_{final}}$} & $\mathrm{d_{final}}$  \\ \hline
{Latent (5Mb)}  & SH     & \multicolumn{1}{c|}{28.0\%} & 0.30           & \multicolumn{1}{c|}{27.4\%} & 0.29 \\ \cline{2-6} 
                & CCE    & \multicolumn{1}{c|}{45.5\%} & 0.18           & \multicolumn{1}{c|}{40.9\%} & 0.22 \\ \cline{2-6} 
                & FCCE   & \multicolumn{1}{c|}{44.7\%} & 0.16           & \multicolumn{1}{c|}{38.7\%} & 0.21 \\ \hline
{Native (50Mb)} & SH     & \multicolumn{1}{c|}{23.5\%} & 0.28           & \multicolumn{1}{c|}{22.3\%} & 0.29 \\ \cline{2-6} 
                & CCE    & \multicolumn{1}{c|}{40.2\%} & 0.19           & \multicolumn{1}{c|}{38.1\%} & 0.23 \\ \cline{2-6} 
                & FCCE   & \multicolumn{1}{c|}{38.9\%} & 0.19           & \multicolumn{1}{c|}{37.5\%} & 0.22 \\ \hline
\end{tabular}
\end{table}

\begin{table}[]
\centering
\caption{Comparison of balancing strategies in the FPT scenario.}
\label{tab:FPT}
\begin{tabular}{|c|c|cc|cc|}
\hline
{strategy}      & {loss} & \multicolumn{2}{c|}{weighted}           & \multicolumn{2}{c|}{not weighted}     \\ \cline{3-6} 
                &        & \multicolumn{1}{c|}{$\mathrm{a_{final}}$} & $\mathrm{d}$  & \multicolumn{1}{c|}{$\mathrm{a_{final}}$} & $\mathrm{d}$  \\ \hline
{Latent (5Mb)}  & SH     & \multicolumn{1}{c|}{25.6\%} & 0.28   & \multicolumn{1}{c|}{25.6\%} & 0.29 \\ \cline{2-6} 
                & CCE    & \multicolumn{1}{c|}{51.6\%} & 0.19   & \multicolumn{1}{c|}{48.6\%} & 0.22 \\ \cline{2-6} 
                & FCCE   & \multicolumn{1}{c|}{48.7\%} & 0.17   & \multicolumn{1}{c|}{47.1\%} & 0.19 \\ \hline
{Native (50Mb)} & SH     & \multicolumn{1}{c|}{48.8\%} & 0.28   & \multicolumn{1}{c|}{48.6\%} & 0.28 \\ \cline{2-6} 
                & CCE    & \multicolumn{1}{c|}{57.5\%} & 0.19   & \multicolumn{1}{c|}{56.5\%} & 0.21 \\ \cline{2-6} 
                & FCCE   & \multicolumn{1}{c|}{57.5\%} & 0.19   & \multicolumn{1}{c|}{54.5\%} & 0.20 \\ \hline
\end{tabular}
\end{table}

\section{Semi-supervised pre-training of the FE}\label{sec:pt}

A common practice for learning feature representation is to perform supervised learning on a pre-training task, with data $D^{PT}$. In this manner, extracted features are dimiscrinative for the pre-training classes, however, there is no guarantee that they are also discriminative for the tasks that are incrementally learned after the pre-training, named downstream tasks. This is particularly problematic for conventional Latent replay, in which the feature extractor is not updated. 
In this section we challenge the common fully supervised pre-training approach, elaborating an approach to learn richer, transferable features. We first discuss Self-Supervised Learning (SSL) as a promising task-agnostic alternative. We then propose an activation regularization term that promotes features diversity. Finally we evaluate the influence of both on the CIFAR50+5x10 dataset. 

\subsubsection{Self-Supervised Regularization} Learning representation with self-supervision can boost performance in CIL settings \cite{protoAug}\cite{pham2021dualnet}. Introducing a proxy task, such as clustering \cite{caron2020unsupervised}, rotation prediction \cite{Gidaris2018UnsupervisedRL}, contrastive loss~\cite{he2020momentum}, or maximizing the information content of
the embedding~\cite{zbontar2021barlow}, makes the feature representation more task-agnostic. SSL thus prevents FE to over-specialize on a given task. 

To validate the interest of SSL for BNN pre-training, we adapt the state-of-the-art Barlow Twin (BT) loss \cite{zbontar2021barlow}. SSL on BNN is a recent research field and early work~\cite{shen2021s2} has highlighted the difficulties to adapt contrastive approaches, \ie, MOCO-V2~\cite{he2020momentum}, without complex distillation loss and multi-phase training. Furthermore, existing results are obtained on larger BNN architecture with BN layers. To be compatible with our compact FBNN with single-phase training we look for self-supervised approach that adds minimal mechanisms. The BT approach achieves state-of-the-art performance with minimal computational overhead. Unlike other methods, the BT has the advantage to work with small batch size, which decreases the chance of training issues in FBNN.

The overall approach is described Fig. \ref{pretraining:loss_schema}. Two augmentations $X_1$ and $X_2$ of the same image batch $X$ are projected with a projector block after FE. We denote as $\mathcal{C}$ the correlation matrix between the two projections. The self-supervision proxy task consists in minimizing the distance between $\mathcal{C}$ and the identity matrix. We can decompose the loss into two terms, as presented in (\ref{eq:bt_loss}). The first one encourages the diagonal components to be close to $1$ enforcing feature invariance to data augmentation and the second one minimizes off-diagonal terms preventing features redundancy. Both terms are weighted with $\lambda = 10^{-5}$, as in the original paper. 

\begin{equation}\label{eq:bt_loss}
    \mathcal{L}_{SSL} = \sum_{i}(1 - \mathcal{C}_{ii})^2 + \lambda \sum_{i} \sum_{j \neq i} \mathcal{C}_{ij}^2
\end{equation}

\subsubsection{Features regularization} Since binary latent features seem highly correlated by classes, we believe that forcing latent features to be more distributed (less correlated) in the binary latent space could help capturing more diverse characteristics. We thus propose a regularization term $\mathcal{L}_{FR}$ that promotes components to have opposite binary value within a batch of samples. As described in (\ref{eq:actReg}), this activation regularization loss acting on $\mathbf{z}$ minimizes the component's sum along the batch axis $B$, for each of the $D$ latent features.

\begin{equation}\label{eq:actReg}
    \mathcal{L}_{FR} = \frac{1}{D\times B} \sum_{d=1}^{D} \sum_{b=1}^{B} z_{b,d}
\end{equation}

\subsubsection{Multi-objective weighted loss} SSL and latent features regularization adds two proxy tasks that require no label, which we believe is key to learn more transferable features. However, during pre-training  labels are available. Moreover, Section~\ref{sec:lb} indicates that that supervised learning with CCE is a good baseline for representation learning. We therefore propose a combination of both supervised and self-supervised learning, and hence the loss minimization is formulated as a multi-objective optimization problem.
$\mathcal{L}_{CCE}$, $\mathcal{L}_{SSL}$ and $\mathcal{L}_{r}$ are then combined in a single loss with $\alpha$, $\beta$ and $\gamma$ weights:

\begin{equation}\label{eq:sup_ssl_reg_loss}
    \mathcal{L}_{PT} =  \alpha \mathcal{L}_{CCE} + \beta \mathcal{L}_{SSL} + \gamma \mathcal{L}_{FR} 
\end{equation}

\begin{figure}[!h]
\centering
\includegraphics[trim=0 300 0 0,clip, scale = 0.5]{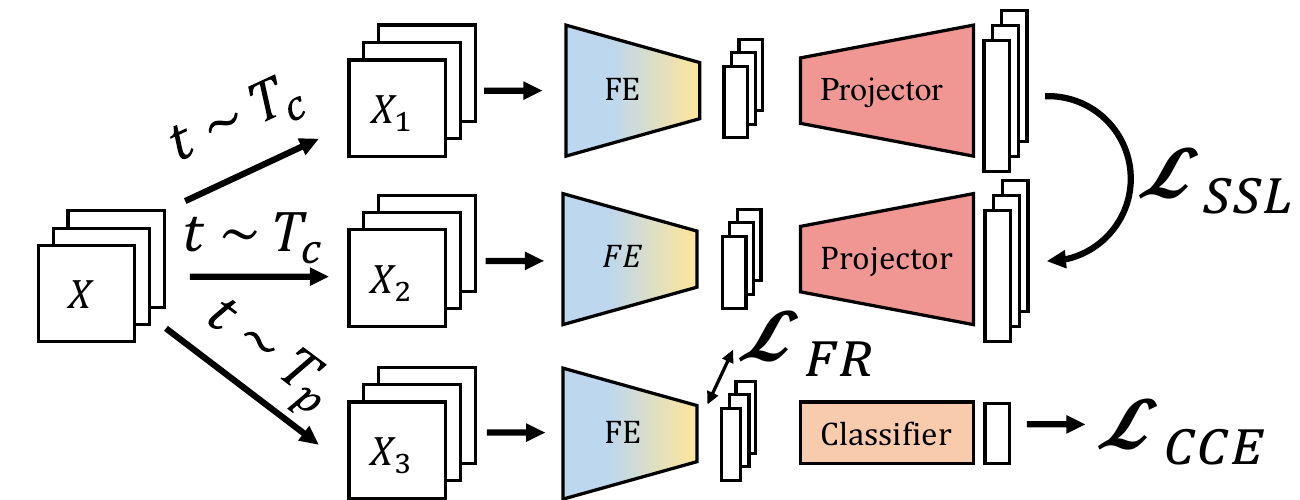}
\caption{Illustration of the multi-objective pre-training framework.}
\label{pretraining:loss_schema}
\end{figure}

The proposed pre-training loss is computed as described in Fig. \ref{pretraining:loss_schema}. $\mathcal{L}_{CCE}$ and $\mathcal{L}_{FR}$ are computed using the previously mentioned data augmentation $T_{p}$ (Section \ref{sec:eval_methodo}) on each input batch $X$. Since $\mathcal{L}_{SSL}$ requires a stronger data augmentation, we thus added random color transformations and random crop, denoted as $T_{c}$. As the pre-training can be done without compute resource constraints, the projection block is composed of the stack of two full-precision dense layers with $2048$ neurons followed by a BN and a ReLU activation.  

\subsection{Experimental Results}

Loss terms compromises are evaluated on \textit{CIFAR50+5X10} for Latent replay, a 10Mb memory buffer, the \textit{3Mb-BNN} model and a class-balanced CCE. Table~\ref{tab:pt_metrics} reports metrics for the end of the pre-training stage and at the end of the 5 retrainings.

\begin{table*}[]
\centering
\caption{Final retraining performance of CIFAR50+5X10 using Latent replay (10Mb buffer) for various regularization setups.}
\label{tab:pt_metrics}
\begin{tabular}{|c|ccc|ccc|ccc|}
\hline
paradigm            & $\alpha$ & $\beta$     & $\gamma$    & $\mathrm{a_{PT}^{train}}$ & $\mathrm{a_{PT}^{test}}$ & $\mathrm{d_{PT}}$ & $\mathrm{a_{final}^{train}}$ & $\mathrm{a_{final}^{test}}$ & $\mathrm{d_{final}}$ \\ \hline
supervised     & 1 & 0         & 0         & 90.74\% & 62.90\% & 0.15 & 56.33\% & 51.28\% & 0.17 \\ \hline
regularization & 1 & 0         & $10^{-3}$ & 90.82\% & 64.06\% & 0.14 & 55.78\% & 51.86\% & 0.18 \\
               & 1 & 0         & $10^{-2}$ & 90.52\% & 63.26\% & 0.14 & 55.58\% & 51.96\% & 0.19 \\
               & 1 & 0         & $10^{-1}$ & 86.85\% & 61.04\% & 0.14 & 51.97\% & 48.62\% & 0.19 \\ \hline
SSL            & 1 & $10^{-6}$ & 0         & 90.09\% & 64.16\% & 0.13 & 56.47\% & 51.08\% & 0.18 \\
               & 1 & $10^{-5}$ & 0         & 89.57\% & 63.58\% & 0.14 & 57.63\% & 51.92\% & 0.18 \\
               & 1 & $10^{-4}$ & 0         & 86.12\% & 60.34\% & 0.16 & 57.88\% & 51.26\% & 0.18 \\
               & 1 & $10^{-3}$ & 0         & 63.84\% & 48.40\% & 0.17 & 50.04\% & 41.76\% & 0.19 \\ \hline
regularization +SSL & 1        & $5.10^{-6}$ & $5.10^{-3}$ & 90.24\%          & 63.54\%         & 0.14     & 56.53\%             & 52.34\%            & 0.20        \\
               & 1 & $10^{-5}$ & $10^{-2}$ & 90.02\% & 63.51\% & 0.14 & 57.43\% & 52.45\% & 0.19 \\ \hline
\end{tabular}
\end{table*}

\subsubsection{The SSL term} The BT loss effectively improves the final performance for $\beta = 10^{-5}$, for both train and test accuracy (+1.3pts and +0.7pts). Hence, the representation learned with SSL is better suited to downstream tasks.
Indeed, SSL eases the convergence on new tasks, boosting generalization on the test set. The SSL term also enhances the test accuracy on the pre-training task, $\mathrm{a_{PT}^{test}}$, while reducing the train accuracy, $\mathrm{a_{PT}^{train}}$ for $\beta \leq 10^{-5}$. This observation indicates that the SSL term forces the representation to be less discriminative for the training set and it helps regularizing the network.

\subsubsection{The latent features regularization} Feature regularization improves both pre-training test accuracy and final test accuracy for $\gamma \leq 10^{-2}$ by 1.16pts and 0.58pts. A too strong regularization impairs both initial and final performances. 

However, contrary to $\mathcal{L}_{SSL}$, $\mathcal{L}_{FR}$ improves the pretraining accuracy and reduces the final accuracy on train set. It indicates that the loss helps the optimization on the pre-training task but not on downstream tasks. This can be alleviated by combining both SSL and feature regularization loss with best performance reached with $\beta = 10^{-5}$ and $\gamma = 10^{-2}$ increasing final test accuracy by 1.17pts. Note that the best configuration $(\alpha,\beta,\gamma)$ tested appears to be close to the optimum reached for $\beta$ and $\gamma$ when used independently. 

\subsubsection{Main takeaways} This experiment underlines the importance of the pre-training quality and its influence on the CIL performance. Alternative learning paradigms, as self-supervised learning are promising solutions to learn more transferable features without considerable additional costs. 

\section{Native vs. Latent replay}\label{sec:es}

This section compares the performance of Native replay and Latent replay for a given memory footprint, comparison also referred to as 'at iso-memory'. 
The qualitative trade-offs between these two approaches are summarized in Table \ref{tab:es}.

In our specific configuration, Latent replay allows to store $13\times$ more samples per Mb than Native.  
Therefore, Latent replay potentially approximates better past distributions, at iso-memory. This advantage becomes particularly prominent as the number of classes increases. Latent replay also offers computational cost reduction, with $90\%$ less parameters to update and less iterations to convergence (see Section \ref{sec:NND}).

However, this Latent replay gain in number of samples in the buffer is obtained at the cost of a decrease in model reconfigurability. The latent representation can not be updated, otherwise the samples in the buffer would not be representative anymore of the image classes. In particular, this problem is referred to in the literature as the aging problem.

Another non-negligible advantage of Native replay is the possibility to augment the data, which is vital for compact networks. Data augmentation is particularly needed on the memory buffer as only a few samples are kept. However, contrary to Native replay, in Latent replay there is no conventional way to do data augmentation from a binary latent space. 

\begin{table}[h]
\centering
\caption{Pros (+) and cons (-) of Native versus Latent replay.}
\label{tab:es}
\begin{tabular}{ccc}
\hline
Native &                         & Latent \\ \hline
-      & \#Sample/Mb             & ++     \\
+      & Information/sample      & -      \\
-      & Computational cost      & +++    \\
+      & Data Augmentation       & --     \\
+      & Pre-training dependency & --     \\ \hline
\end{tabular}
\end{table}

\begin{figure}[h]
\centering
\includegraphics[trim=0 170 0 0,clip, scale = 0.5]{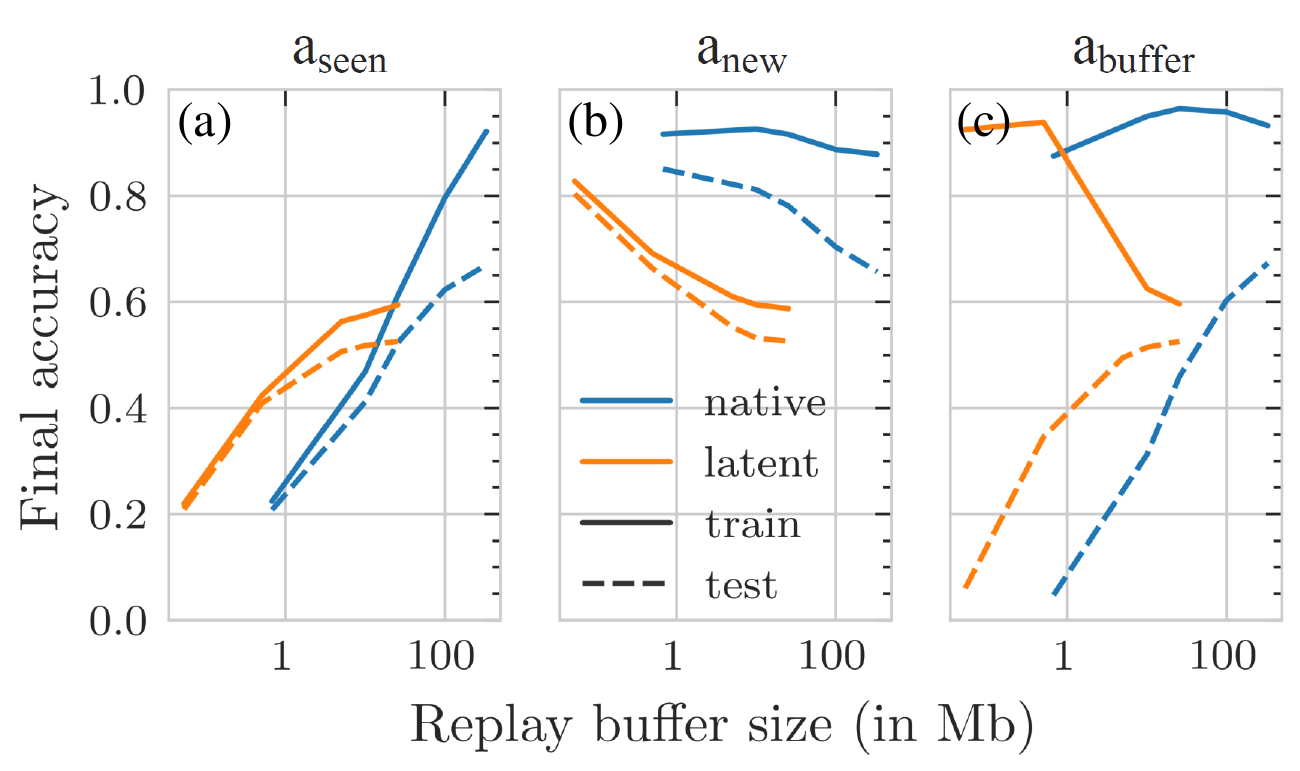}
\caption{Iso-memory comparison between Native and Latent replay.}
\label{fig:perf_memorySize}
\end{figure}

Fig.~\ref{fig:perf_memorySize} presents the impact of the replay buffer size on the final performance for Native and Latent replay with \textit{CIFAR50+5X10}: (a) $\mathrm{a_{seen}}$ represents the final performance on all seen classes, (b) $\mathrm{a_{new}}$ represents the adaptation capacity, \ie, on new training data (without buffer samples), and (c) $\mathrm{a_{buffer}}$ represents the retention capacity, \ie, on past classes.  
The previously presented loss balancing and pre-training are applied. To exhibit the influence of the buffer size on the FE training policy, the Latent and Native strategies are evaluated for buffer sizes ranging from $0$ to the entire dataset.

\subsubsection{Final performance (a)} We observe that, under $25$Mb, Latent replay offers the best performance. Above $25$Mb, Native replay gives better performance. One can note that the $25$Mb point corresponds to the upper bound for latent memory buffer, \ie, all the training dataset can be stored, it brings no improvement to Latent replay to have a larger memory size.

\subsubsection{Adaptation (b)} When we consider the performance on new classes, \ie, $\mathrm{a_{new}}$ accuracy, two distinct interpretations of replay regularization can be made. For latent replay, the augmentation of $\mathrm{m_b}$ decrease both train and test performance by the same magnitude (approximately $30\%$ from 1 sample/image to full dataset). Optimization on new task gets harder. 
On the contrary, for Native replay only the test accuracy decreases. The FBNN still has the ability to model the new training distribution but it has lost its ability to generalize on it. 

\subsubsection{Retention (c)} When looking at the test accuracy on old classes, \ie, dotted lines, we first note, as expected, a sharp increase in retention for larger buffer size for both strategy. However, when the FE is non trainable, the performance decreases on a larger buffer, highlighting the limited modeling capacity. On the contrary, a trainable FE maintains high accuracy across the buffer size, on buffer samples. We nevertheless observe a slight decrease for small buffers, likely explained by the strong imbalance in the training set $D^t \cup M$.

\subsubsection{Main takeaways} The choice to let the FE trainable depends on the memory budget. For a small memory buffer ($<$25Mb), Latent replay yields higher final accuracy than Native. In this case, Latent offers a better retention but lower adaptation than Native. If obtaining the best performance on the current task at the cost of less retention is preferred, the FE can be trainable, regardless of the buffer size. 

\section{Evaluation on the CORE50 dataset}\label{sec:val}

In this section, we evaluate our approach developed in the previous sections with a more realistic benchmark, the \textit{CORE50} dataset~\cite{core50}.  
Recently introduced in the CIL community, \textit{CORE50} serves to benchmark strategies in real-life scenarios. \textit{CORE50} comprises $128\times128$ pixels images belonging to 50 classes of objects acquired across 11 recording sessions, with 300 consecutive frames (images) per session. 

We adopt the scenario of \cite{pham2021dualnet} to compare our approach to prior work. It consists in 10 retraining tasks of 5 classes obtained by randomly splitting the dataset. Sessions \#3,\#7, and \#10 represent the test dataset and the others the train dataset. 
Compared to \textit{CIFAR100}, \textit{CORE50} enables us to extend our conclusions to larger, more complex images, with temporal correlations during acquisition and under various backgrounds.

A deeper architecture, \textit{3Mb-Res-BNN}, is designed to comply with \textit{CORE50} requirements, adhering to the previously stated principles and remaining within the 3Mb model memory size budget. 
We subsequently evaluate \textit{3Mb-Res-BNN} Latent and Native replay against existing approaches.
As there is no pre-training task in the \textit{CORE50} scenario, we propose to pre-train our \textit{3Mb-Res-BNN} model on the \textit{STL10} dataset~\cite{coates2011analysis}.

This is a well-know benchmark for evaluating unsupervised and semi-supervised learning algorithms. It consists of 10 classes, including various object categories. 
The dataset features 5000 labeled training $96\times96$ color images (500 images/class) and 100000 unlabeled images. 

\subsection{Deeper FE with skip-connections}\label{subsec:core50bnn} 
\begin{figure}[!h]
\centering
\includegraphics[trim=0 90 0 0,clip, scale = 0.5]{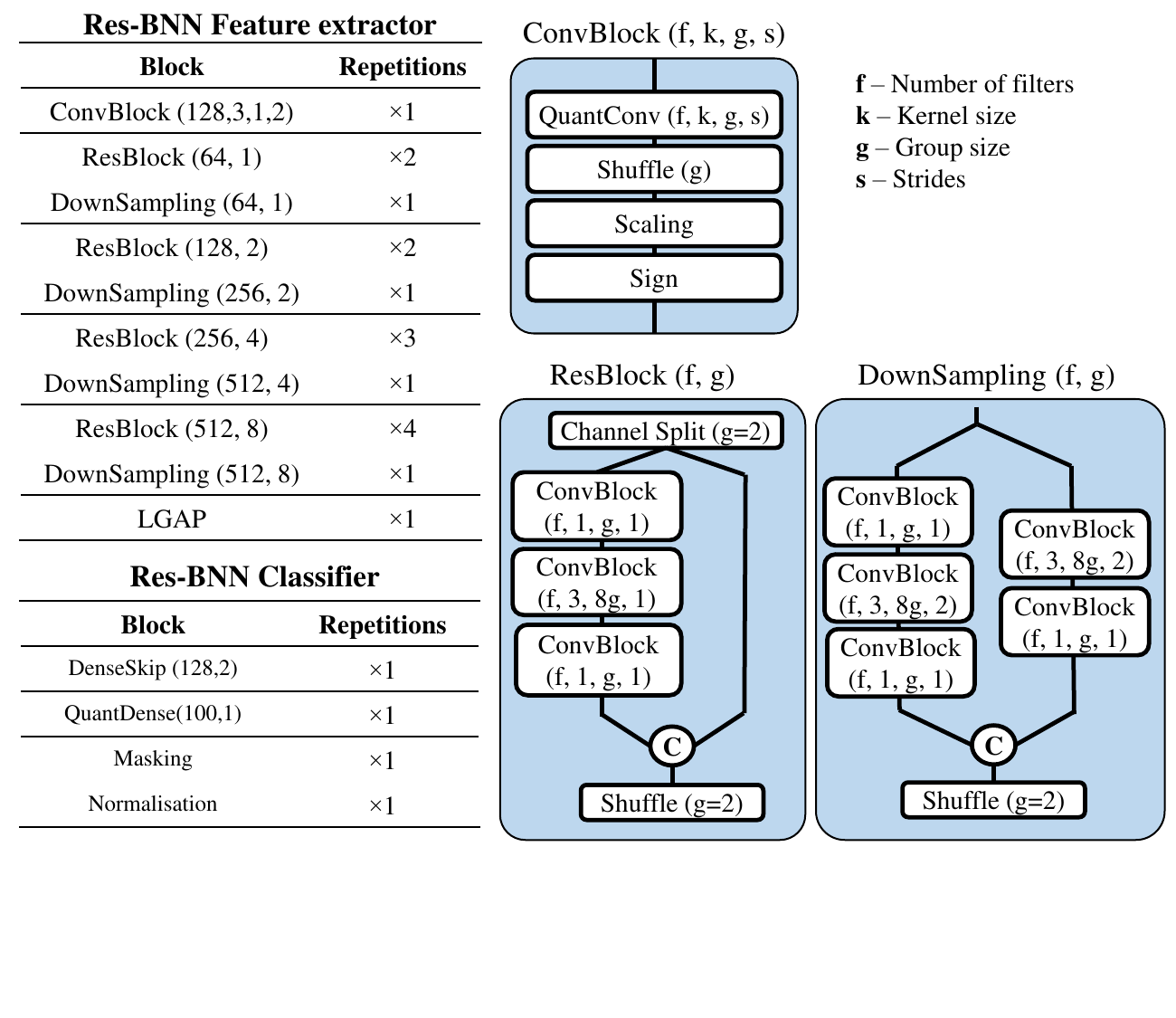}
\caption{\textit{3Mb-Res-BNN} architecture - Deeper FE to comply with CORE50.}
\label{fig:NewArchitecture}
\end{figure}

\textit{3Mb-Res-BNN} accommodates larger images thanks to a deeper architecture composed of residual blocks, as depicted in Fig.~\ref{fig:NewArchitecture}. After a first down-scaling with a convolution block of stride 2, we stacked residual blocks that preserve the spacial feature size, and down-sampling blocks that halve the spatial feature size, inspired by ShuffleNetv2~\cite{Shufflenet}. For the residual block, the main and residual inputs paths are obtained by splitting the feature map in half in the channel dimension. The main path consists in a stack of $1\times1$, $3\times3$, and $1\times1$ convolution blocks with a higher number of groups on the $3\times3$ convolutions. The down-sampling blocks are similar to the res-blocks but with a stride of on $3\times3$ convolutions. The residual path contains a $3\times3$ and $1\times1$ convolution block. In both blocks, the path aggregation is done by concatenation, as opposed to addition which could lead to bit overflow. After, feature maps are shuffled with a shuffle layer of group 2. 

When putting it all together, \textit{3Mb-Res-BNN} results in a 3M parameters architecture with 17 blocks and a $1024$ dimensions latent space. 
With this FE, Latent replay therefore allows to store $80\times$ more examples than Native, at iso-memory. 

\subsection{Experimental Results}

\begin{table}[h]
\caption{Evaluation metrics on CORE50 Benchmark. A buffer of 50 samples is used for TA and 100 samples for TF.}
\label{tab:CORE50_benchmark}
\centering
\begin{tabular}{lcc}
\hline
Method                         & CORE50-TA        & CORE50-TF        \\ \hline
ER~\cite{Ratcliff1990} (reported from \cite{pham2021dualnet}) & 41.72 $\pm$ 1.30 & 21.80 $\pm$ 0.70 \\
ER-Aug                         & 44.16 $\pm$ 2.05 & 25.34 $\pm$ 0.74 \\
DER++~\cite{buzzega2020dark} (reported from \cite{pham2021dualnet})   & 46.62 $\pm$ 0.46 & 22.84 $\pm$ 0.84 \\
DER++-Aug                       & 45.12 $\pm$ 0.68 & 28.10 $\pm$ 0.80 \\ \hline
CTN~\cite{pham2021contextual} (reported from \cite{pham2021dualnet})  & 54.17 $\pm$ 0.85 & N/A              \\
CTN-Aug                        & 53.40 $\pm$ 1.37 & N/A              \\
DualNet~\cite{pham2021dualnet} & 57.64 $\pm$ 1.36 & 38.76 $\pm$ 1.52 \\ \hline
\textit{3Mb-Res-BNN} Latent replay (ours)                 & \textbf{53.96} $\pm$ 1.17 & \textbf{17.22} $\pm$ 0.16 \\
\textit{3Mb-Res-BNN} Native replay (ours)                 & \textbf{59.07} $\pm$ 1.24 & \textbf{44.52} $\pm$ 0.54 \\ \hline
\end{tabular}
\end{table}

\begin{figure}[h]
\centering
\includegraphics[trim=0 180 0 0,clip, scale = 0.5]{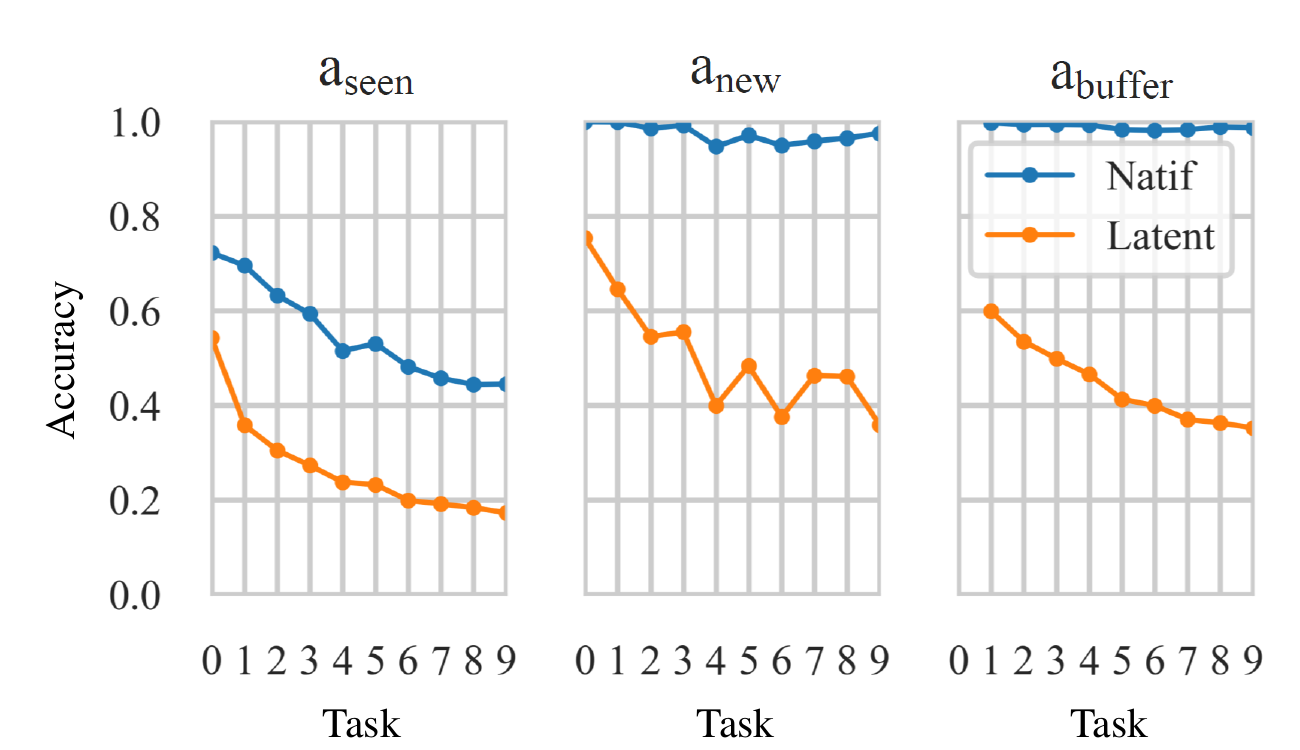}
\caption{Task-Free (TF) CIL performance for \textit{3Mb-Res-BNN} in Native and Latent replay on \textit{seen},\textit{new} and \textit{buffer} subset. 100 samples per class are stored for Native replay and we use an equivalent buffer size for Latent replay.} 
\label{fig:val_IncrementalLearning}
\end{figure}

The STL-10 pre-training consists of a multi-phase training protocol with BN, to start from a better weight initialization. 
The first phase consists in a standard supervised training only on labeled images only. BN layers are then replaced by scaling-factors, as introduced in Section~\ref{sec:key_cil}. 
Then, a second training phase is performed on the entire dataset, benefiting from unlabelled data for the unsupervised training loss terms. $\mathcal{L}_{CCE}$ is computed on labeled images and $\mathcal{L}_{FR}$ and $\mathcal{L}_{SSL}$ on unlabelled images, for regularization. The training hyperparameters are kept the same as in the previous sections. 

The comparison to prior work is rather difficult, as few approaches investigate exactly the same CIL scenario. Nevertheless, we compare our approach to continual learning approaches with a replay buffer \cite{pham2021dualnet}. 
The baselines are ER~\cite{Ratcliff1990}, DER++~\cite{buzzega2020dark}, an ER variant with knowledge distillation on logits, CTN~\cite{pham2021contextual} a TIL approach with fixed FE and DualNet~\cite{pham2021dualnet} a CIL method that uses SSL to slowly update the FE. The first three methods are evaluated (Table \ref{tab:CORE50_benchmark}) with or without data augmentation (-Aug suffix). 
The network used in these articles relies on ResNet18~\cite{he2016deep} with a 352Mb memory size, \ie, more than $100\times$ larger than \textit{3Mb-Res-BNN}. 
Unlike our approach, in~\cite{pham2021dualnet} the CIL is optimized on 3 epochs with a very high learning rate of $0.03$. 
However, \textit{3Mb-Res-BNN} requires smaller learning rate, \ie, $10^{-4}$, (as empirically observed, higher learning rates, $>10^{-3}$, may cause training instability) and hence a larger number of epochs (approximately a hundred) is necessary for convergence.

Table~\ref{tab:CORE50_benchmark} reports the final accuracy as proposed in the literature, without task-label (Task-Free, TF) and with task-label (Task-Aware, TA), respectively corresponding to CIL and TIL.
We report the final training accuracy for Native replay with 100 images per classes for TF setting and 50 per classes for TA setting, similarly to the other methods in ~\cite{pham2021dualnet}.
In the same manner, we report accuracy with Latent replay with an equivalent memory buffer size.

In the TF case, \textit{3Mb-Res-BNN} with Latent replay is slightly below the range of performance of the state-of-the-art ER methods without relying on any floating point operation and with $100\times$ less parameters. Nevertheless, in TA setting, \textit{3Mb-Res-BNN} Latent replay outperforms ER, DER and CTN with its $53.96\%$ final accuracy.
\textit{3Mb-Res-BNN} with Native replay presents state-of-the-art performance in TF with $44.52\%$, which is far better than DualNet, despite of using a $100\times$ smaller model memory footprint, a $7.4\times$ smaller buffer footprint thanks to our $TYCC-16$ image encoding and binary-only arithmetic that tremendously reduces the associated computational complexity. 

We further investigate incremental training dynamics in Fig.~\ref{fig:val_IncrementalLearning} (TF). With Native replay, \textit{3Mb-Res-BNN} succeeds to maintain a high level of adaptation with more than $90\%$ on new task training sets. It also fully captures the knowledge in the buffer with perfect classification score on buffer samples. On the contrary, with Latent replay, \textit{3Mb-Res-BNN}'s ability to fit the training data (new task and buffer) diminishes through retraining from $70\%$ to $37\%$. The classifier alone has not the capacity to fit the whole classification problem. However, we can conclude that the FE can still extract meaningful features while having been trained on a completely different dataset. Indeed on the first CIL task, Latent replay is already $12\%$pts below Native replay while having a fixed FE. 

\section{Conclusion}\label{sec:conclusion}

This study presents an exploration of Fully Binarized Neural Networks (FBNNs) within the context of Class Incremental Learning (CIL), particularly through the lens of Experience Replay (ER). By addressing critical challenges such as network design, loss balancing, and pre-training, we have demonstrated the viability of FBNNs in dynamic environments where computing resources are constrained. Our findings emphasize the importance of tailored network architectures for BNN-CIL, which are crucial for maintaining expressiveness, mitigating convergence issues and keeping a small memory buffer. The integration of carefully designed loss functions and pre-training strategies enhances adaptability and reduces catastrophic forgetting. Furthermore, our experiments highlight the trade-offs between Native and Latent replay methods, with Latent replay proving to be more effective in memory-constrained settings. 
Finally, our CORE50 results shows that for a challenging benchmark, a careful design of a high-end FBNN model (\textit{3Mb-Res-BNN}) as well as its incremental training protocol, can even outperform other CIL frameworks that rely on full-precision network models.

\section*{Acknowledgments}
This work is part of the IPCEI Microelectronics and Connectivity and was supported by the French Public Authorities within the frame of France 2030.


\bibliographystyle{unsrt}

\end{document}